 \documentclass[5p,times]{elsarticle}

\usepackage{algorithm}
\usepackage[noend]{algpseudocode}
\usepackage{tabularx,ragged2e,booktabs}
\usepackage{tabulary}
\usepackage{bm}
\usepackage{soul}
\usepackage{hyphenat}

\usepackage{graphics} 
\usepackage{epsfig} 
\usepackage{mathtools}
\usepackage{times} 
\usepackage{xcolor}
\usepackage{multicol}
\usepackage{multirow}
\usepackage{colortbl}
\usepackage{lipsum}
\usepackage{url}

\DeclarePairedDelimiter{\norm}{\lVert}{\rVert} 

\usepackage[numbers]{natbib}

\journal{ }

\begin{document}

\begin{sloppypar}

\begin{frontmatter}



\title{SUPER-MAN: SUPERnumerary Robotic Bodies for Physical Assistance in HuMAN-Robot Conjoined Actions}


\author[label1]{Alberto Giammarino}
\author[label1]{Juan M. Gandarias\corref{cor1}}
\ead{juan.gandarias@iit.it}
\ead[url]{https://www.iit.it/it/web/hrii/}
\cortext[cor1]{Corresponding author}
\fntext[label1]{Human-Robot Interfaces and Interaction, Istituto Italiano di Tecnologia.\\
Via S. Quirico, 19d. 16163 Genoa, Italy.
}
\author[label1]{Pietro Balatti}
\author[label1]{Mattia Leonori}
\author[label1]{Marta Lorenzini}
\author[label1]{Arash Ajoudani}

\begin{abstract}
This paper presents a mobile supernumerary robotic approach to physical assistance in human-robot conjoined actions. The study starts with the description of the SUPER-MAN concept. The idea is to develop and utilize mobile collaborative systems that can follow human loco-manipulation commands to perform industrial tasks through three main components: i) an admittance-type interface, ii) a human-robot interaction controller and iii) a supernumerary robotic body. Next, we present two possible implementations within the framework – from theoretical and hardware perspectives. The first system is called MOCA-MAN, and is composed of a redundant torque-controlled robotic arm and an omni-directional mobile platform. The second one is called Kairos-MAN, formed by a high-payload 6-DoF velocity-controlled robotic arm and an omni-directional mobile platform. The systems share the same admittance interface, through which user wrenches are translated to loco-manipulation commands, generated by whole-body controllers of each system. Besides, a thorough user-study with multiple and cross-gender subjects is presented to reveal the quantitative performance of the two systems in effort demanding and dexterous tasks. Moreover, we provide qualitative results from the NASA-TLX questionnaire to demonstrate the SUPER-MAN approach's potential and its acceptability from the users' viewpoint.
\end{abstract}




\end{frontmatter}


\section{Introduction}
\label{sec:intro}

Robotic technologies are increasingly present in industrial environments to replace human labor in several repetitive and simple operations~\cite{graetz2018robots}, and to prepare a response to an aging workforce~\cite{krueger2017have}. 
Despite that, robots are still far from achieving human intelligence and versatility, limiting the range of tasks they can perform autonomously~\cite{lake2017building}. Consequently, workers are still involved in physically demanding tasks that might compromise workers' health and productivity~\cite{villani2018survey}. A prime consequence of this fact is that musculoskeletal disorders (MSDs) remain the most common work-related health problem in the European Union (EU)~\cite{dework}. According to that study, the most common types of MSDs are backache and muscular pains in the shoulders, neck, and upper limbs, reported by 46\% and 43\% of the EU workers in 2015, respectively. The physical risk factors mainly related to MSDs are wrong postures, heavy physical work, lifting, repetitive work, and exposure to vibrations from hand tools.

\begin{figure}
    \centering
    \includegraphics[width=0.9\columnwidth]{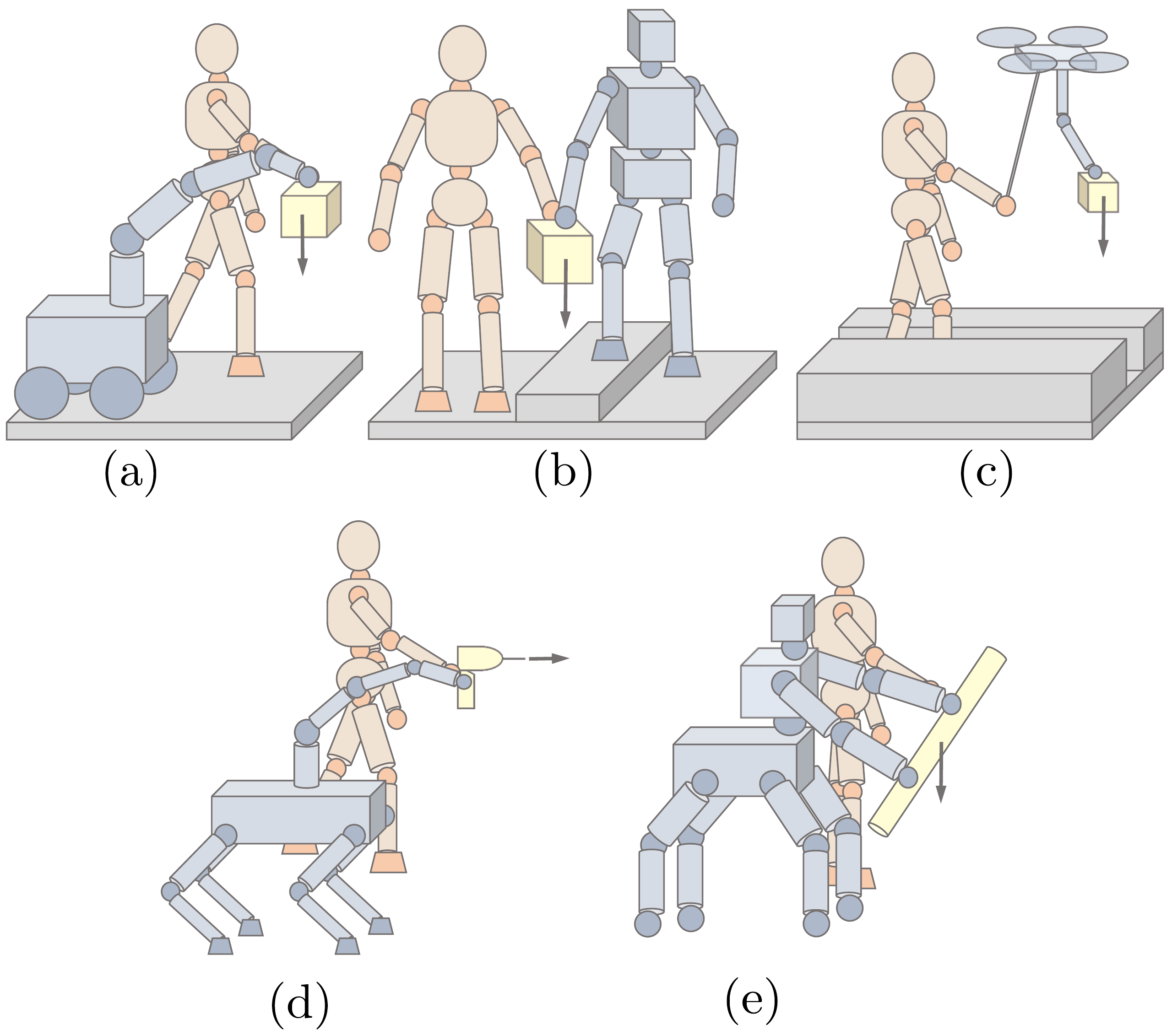}
    \caption{SUPER-MAN framework concept: A physical assistance system involving floating-base robots in human-robot conjoined loco-manipulation tasks. Different robotic platforms can be considered such as (a) wheeled mobile manipulator, (b) humanoids robots, (c) aerial manipulators, (d) legged mobile manipulators, or (e) hybrid (legged and wheeled) robotic systems.}
    \label{fig:mainfigure}
\end{figure}


One viable approach is to build robotic technologies that can work in synergy with a human partner to merge the knacks of both agents. Of particular interest for this paper is the concept of human augmentation~\cite{kazerooni1990human, bougrinat2019design}, which considers a robotic device as physical assistance system, while the human is the central brain of the cooperating dyad. The current research panorama distinguishes three possible macro~-solutions of this type: Exoskeletons, Supernumerary Robotic Limbs~(SRLs), and Collaborative Robots, also known as Cobots.

Exoskeletons are devices humans can wear to enhance their power and assist them in physically demanding tasks.
Although they improve human performance in terms of the effort of the assisted part of the body, several challenges still need to be overcome. Since they are wearable devices, they should be able to accommodate a variety of workers (i.e., adapt to different body geometries and dimensions). The ease of putting them on/off also plays an essential role in the acceptance of this technology~\cite{toxiri2019back}. 
Regarding comfort, the weight of the device must be carried by the user, which might result in overloading other joints/muscles of the body~\cite{de2016exoskeletons, bar2021influence}. Besides, they need to apply pressure on the body to function~\cite{de2016exoskeletons}, and the user may experience discomfort and pressure-related tissue injuries if the contact areas are not carefully designed~\cite{kermavnar2021effects}. They can also hinder the natural movements of the wearer, reducing their mobility and affecting their postural balance~\cite{theurel2019occupational, kermavnar2021effects}.

More recently, SRLs were introduced to deal with the issues and challenges mentioned before.  Unlike exoskeletons, SRLs provide additional limbs to extend and enhance users' capabilities. 
Still SRLs share with exoskeletons most of the challenges related to their wearable nature, e.g., the device's weight and discomfort due to workers' reduced mobility~\cite{tong2021review}. Moreover, SRLs also present some unique challenges, e.g., they must compensate for interferences arising from the wearer body's motion if they want to provide robot-like accuracy~\cite{tong2021review}. Today, essential aspects like wearability, efficiency, and usability are still to be compromised for these technologies~\cite{yang2021supernumerary}. Another controversial issue with the SRLs and exoskeletons is that, they function only when coupled with humans. This means that, they cannot be used independently to perform any (even the simplest) industrial tasks.

Although the concept of Cobot typically refers to fixed-base manipulators, it comes from the Northwestern patent~\cite{colgate1999cobots}, which considers the general case of collaborative robotic systems for direct physical interaction with humans. Here, we use the generalized term to refer to floating-base robotics systems~\cite{wu2019teleoperation}.
According to~\cite{villani2018survey}, Cobots are mainly used to perform tedious tasks, such as moving materials, holding heavy objects, or performing sample tests. Although Cobots have the prospect to overcome some problems presented in exoskeletons and SRLs, they are primarily underused since they are mainly regarded as third-party tools that relieve workers from specific physical tasks.

 This article contributes to the cobots domain when direct physical contact between the robot and the human is required. We present the SUPER-MAN framework, which considers comprehensive supernumerary robotic bodies for physical assistance in human-robot conjoined loco-manipulation tasks (see Fig.~\ref{fig:mainfigure}). This concept builds on our previous work where we presented MOCA-MAN~\cite{kim2020moca}, expanding it to a generic floating-base robotic platform. The general concept of the framework is explained in detail and can be applied to any floating-base platform with loco-manipulation capabilities. Besides, the framework is evaluated through different experiments for two particular supernumerary robotic bodies. Moreover, a user study for each platform is carried out. Hence, the contributions of this work can be summarized as follows:

\begin{itemize}
    \item The proposal of the SUPER-MAN framework as an extension of the MOCA-MAN concept to any generic supernumerary robotic body. The framework principles are described and presented through two wheeled mobile manipulators with different characteristics: MOCA and Kairos. The framework integration in these platforms is respectively called MOCA-MAN and Kairos-MAN.
    \item A usability study with twelve human subjects performing two different industrial-like activities with and without robotic assistance. Each task is selected to suit the characteristics of the corresponding robotic platform used for human assistance. Quantitative and qualitative results are analyzed in order to assess the usability of SUPER-MAN. In particular, the subjects undergo the NASA-TLX questionnaire~\cite{hart1988development}, and task-related quantitative performances are measured to evaluate usability from different points of view.
    \item Further experimentation shows two contrasting cases in which one platform is preferred over the other. A peg-in-hole experiment is carried out to show the excellent interaction capabilities of MOCA-MAN. On the other hand, a high-payload pick-and-place experiment exhibits the good cargo capabilities of Kairos-MAN.
\end{itemize}

The remainder of the paper is organized as follows.
Section~\ref{sec:related_work} presents the relevant literature for the problem considered in this work. Then, Section~\ref{sec:principles} describes the primary principles of the SUPER-MAN framework and develops and demonstrates two possible implementations, i.e. MOCA-MAN and Kairos-MAN. Later, Section~\ref{sec:experiments} shows experiments and results conducted in this work. Finally, Section~\ref{sec:discussion} discusses the experimental results and highlights advantages and drawbacks of the framework proposed and Section~\ref{sec:conclusions} draws the conclusions.

\section{Related Work}
\label{sec:related_work}

This section provides a literature review of the works embraced by the research topic of collaborative robotics for human augmentation. In particular, existing solutions in the three main macro-areas previously mentioned (i.e., Exoskeletons, SRLs, and Cobots) are presented.

\subsection{Exoskeletons}
\label{subsec:exoskeletons}

Exoskeletons are usually classified based on three main characteristics: the presence or not of actuators (i.e., active or passive), the affinity with the human body (i.e., anthropomorphic or non-anthropomorphic), and the assisted part of the body (i.e., upper-body, lower-body, or full-body)~\cite{de2016exoskeletons}. 
For industrial environments, occupational exoskeletons can be divided into back support and upper limb exoskeletons, where the most common applications aim at preventing Low Back Pain (LBP) and shoulder MSDs~\cite{theurel2019occupational}.
Overhead works are frequent in some industries, and many of the related works on upper limb exoskeletons focus on these types of tasks, while the use of these devices can be relevant also for other operations like manual handling~\cite{theurel2018physiological}. Some of the most relevant findings of these works are listed below:

\begin{itemize}
\item The importance of evaluating the designs under three dimensions of potential outcomes: physical demand, task performance, and usability~\cite{alabdulkarim2019influences}, and how mechanical loads might be shifted or transferred~\cite{weston2018biomechanical}. 
\item The high effectiveness and acceptance of a passive device in overhead activities~\cite{huysamen2018evaluation}. 
\item The level of support provided by the device might affect its acceptance, and the optimal level of support is person-dependent in both active and passive exoskeletons~\cite{sylla2014ergonomic, van2019experimental}.
\item An exoskeleton can substantially reduce upper extremities physical demand, but it has relatively small adverse effects on low back physical demand and discomfort~\cite{rashedi2014ergonomic, alabdulkarim2019influences}. 
\item The benefits of the exoskeleton to reduce shoulder flexor muscle activity are demonstrated in ~\cite{theurel2018physiological}. However, broader physiological consequences have also been evidenced as increased antagonist muscle activity, postural strains, cardiovascular demand, and modified upper limb kinematics. 
\item It remains unclear if unexpected health benefits or concerns are due to differences in exoskeleton design or to characteristics of work tasks considered~\cite{kim2018assessingI, kim2018assessingII}.
\end{itemize}

\begin{figure*}
    \centering
    \includegraphics[width=0.9\textwidth]{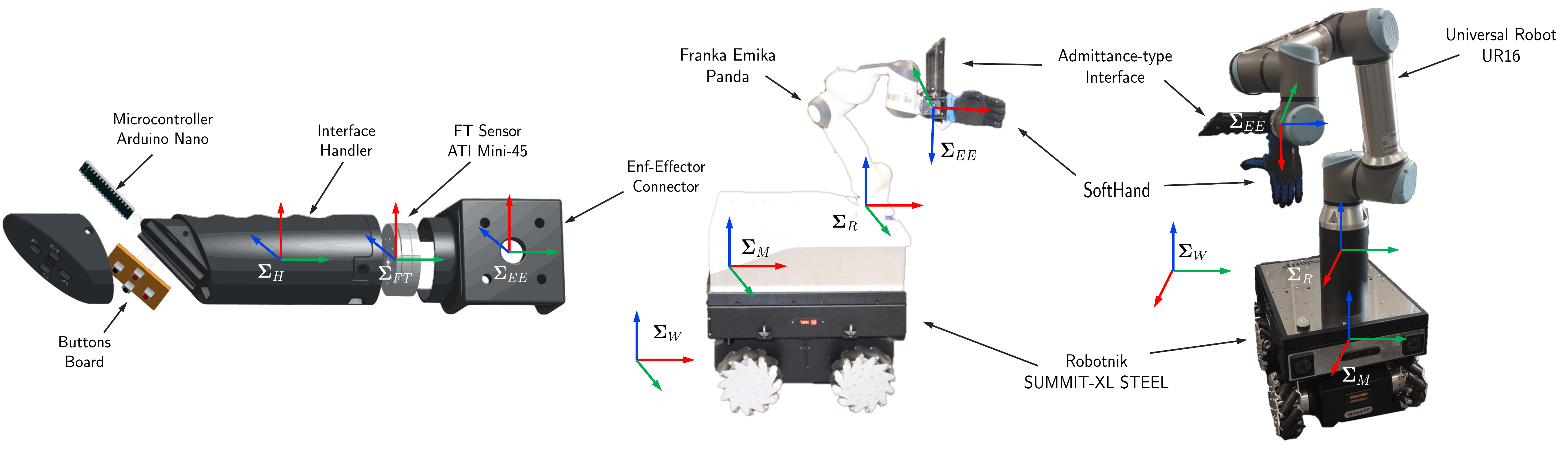}\\
     \hspace{0.7cm} (a) \hspace{4cm} (b) \hspace{4.3cm} (c)
    \caption{Systems overview: (a) Admittance-type Interface. The interface used in this work features a simplistic design with a Force/Torque Sensor for the admittance controller (see section~\ref{subsubsec:admittance_controller}) and a button board to communicate with the robot. Two robotic platforms are considered as supernumerary bodies: (b) MOCA and (c) Kairos. Both platforms integrate an omni-directional base but a different manipulator with diverse characteristics.}
    \label{fig:systems}
\end{figure*}

\subsection{Supernumerary Robotic Limbs}
\label{subsec:supernumerary_limbs}

SRLs are usually classified as Supernumerary Robotic Arms (SRAs), Supernumerary Robotic Legs (SRLGs), and Supernumerary Robotic Fingers (SRFs) depending on which part of the body they augment~\cite{yang2021supernumerary}. Here we consider the SRLs in industrial applications, predominantly SRAs and SRLGs.

In~\cite{gonzalez2018design}, an SRLG system named Extra Robotic Legs (XRL) is developed to assist workers who wear Personal Protection Equipment (PPE) while walking and assuming ergonomically complex postures, e.g., kneeling and crouching for near-ground tasks. One of the challenges is how to compromise conflicting requirements of the actuators to have a satisfying load-bearing capacity for supporting workers' and equipment's weight and high bandwidth to better control the dynamic interactions with the ground. 
Then, in~\cite{parietti2016supernumerary} an SRA system is designed and tested for supporting the human against elements of the surrounding environment while performing a task. 

As previously mentioned, overhead works are one of the most frequent and problematic tasks in the industry. An SRA technology for overhead works is presented in~\cite{bonilla2014robot}. The system is secured to the human's shoulders, and an algorithm is implemented and tested for proactive collaboration with the human in a ceiling panel installation task. Although all the previous works and others like~\cite{davenport2012design, parietti2013dynamic} present successful implementations of design and control techniques for SRLs, they lack in thorough user studies which are of fundamental importance for determining the acceptance of such technologies from the worker point of view.

\subsection{Cobots}
\label{subsec:cobots}

Although cobots are not needed to be worn and tend to be multipurpose, the number of applications in which intentional physical contact between the cobot and the human occurs is minimal. Rehabilitation applications are one of the most common fields in which cobots are starting to be used in that sense. In this regard, cobots are used under therapists' supervision for safety reasons~\cite{westerveld2014damper}. However, with the improvements in security concerns, recent solutions in medical robotics are considering more autonomous approaches~\cite{gandarias2019underactuated}. In~\cite{ruiz2021upper}, an autonomous method for the estimation of the kinematic parameters of the human arm is presented. In that work, a fixed-base cobot with a Cartesian impedance controller follows a compliant trajectory according to human arm kinodynamics to get proprioceptive data for estimation. In~\cite{ding2022intelligent}, an admittance-controlled mobile manipulator is used to give support to older adults while walking.

Regarding industrial applications, to the best of our knowledge, our past work~\cite{kim2020moca} is the only one in which a mobile cobot is used to provide support to a worker with direct physical contact, similarly to exoskeletons or SRLs approaches. The system considered an EMG-based bracelet to command the robot with one hand and a detachable clamp to attach to the manipulator with the other hand.

\section{SUPER-MAN Principles}
\label{sec:principles}

This section describes the main features of the SUPER-MAN framework, which is composed of three basic elements: i) an Admittance-type Interface for human-robot coupling, ii) a Human-Robot (HR) Interaction Controller to allow the user to command the robot and iii) a Supernumerary Body (i.e., a floating-base robot) with its Whole-Body Controller that gives physical assistance to the user. These three elements and their characteristics are described below in sections~\ref{subsec:interface},~\ref{subsec:interaction_controller}, and~\ref{subsec:supernumerary_bodies}, respectively. 
For the sake of clarity, from now on we refer to the combination of \textit{mobile platform} and \textit{manipulator} as a \textit{robot} in general, or as a \textit{mobile manipulator}. The manipulator will be referred to as \textit{arm} or \textit{robotic arm}. The \textit{mobile base} will be referred to as the \textit{mobile platform} or the \textit{base}.

In this work, we consider two possible implementations of the SUPER-MAN framework by targeting dexterity and interaction capacity in one, and the effort compensation capacity in the other. These two implementations share a similar admittance-type interface and its HR controller, and differ in terms of hardware and software for the supernumerary bodies.

\subsection{Admittance-type Interface}
\label{subsec:interface}
The SUPER-MAN framework needs an admittance interface that permits the coupling between the human and the robot to assist some parts of the human body based on the task requirements. This interface has to be designed to allow the human to operate the robot locally without compromising the motion of the human body parts not involved in the task. This way, different interfaces may be considered. In our first work related to the SUPER-MAN framework~\cite{kim2020moca} we employed a detachable wristband with a magnetic-based clamp for coupling and a Force/Torque (F/T) sensor, and an EMG-based wristband for commanding the robot.

Conversely, the interface used in this work is shown in Fig.~\ref{fig:systems}a. It consists of two main components: i)~an Arduino Nano microcontroller connected to a button panel that allows the user to configure different functionalities and communicate with the robot through the Robot Operating System (ROS) middleware suite and ii)~a F/T sensor to measure the user interaction wrenches. Overall, the interface exhibits the following features:

\begin{itemize}
    \item A simplistic design that promotes usability and eases the human-robot coupling. As a result, the whole SUPER-MAN system can be operated locally using only one hand.
    \item User-centered capabilities: the worker can control different robot functionalities online based on the need of the task and user preferences.
    \item The interface is programmable and configurable, allowing for better flexibility among the SUPER-MAN functionalities. In this work, the following functionalities are integrated within the button panel: i)~A-button ($\mathcal{A}$): activates/deactivates the admittance controller (see section~\ref{subsubsec:admittance_controller}), ii)~M-button ($\mathcal{M}$): changes the motion mode between translation and roto-translation, iii)~G-button ($\mathcal{G}$): closes/opens the gripper, iv)~P-button ($\mathcal{P}$): switches the priority of motion between the base and the arm for loco-manipulation tasks. A brief explanation of these priorities is included below.
\end{itemize}

\subsection{Human-Robot Interaction Controller}
\label{subsec:interaction_controller}
The SUPER-MAN framework needs to include a high-level controller that allows interactive communication between humans and robots. This controller has to map human intentions onto robot motions or behaviors, allowing haptic guidance and shared control. 

\subsubsection{Admittance Controller}
\label{subsubsec:admittance_controller}
This work implements an admittance controller to read user forces (i.e., user intentions) and send motion references to the robot.
Hence, through the proposed admittance interface, the user commands the desired end-effector motion thanks to the combination of the integrated F/T sensor and the following standard admittance control law
\begin{equation}
    \boldsymbol{M}_{adm}\boldsymbol{\ddot{x}}_d+\boldsymbol{D}_{adm}\boldsymbol{\dot{x}}_d=\hat{\boldsymbol{\lambda}}_h.
    \label{eq:admitcontrollaw}
\end{equation}
All the mathematical symbols are defined in~\ref{appendix:math_not_symb}. Equation~\ref{eq:admitcontrollaw} is solved for desired twist $\boldsymbol{\dot{x}}_d$ and pose $\boldsymbol{x}_d$. Then, the desired motion is sent to the whole-body controller of the robot. As previously mentioned, the 4-buttons board gives the users control over the admittance control law, allowing them to change the motion mode (M-button) between translation and roto-translation and activate/deactivate the controller (A-button).

\subsubsection{Loco-Manipulation Skills}
\label{subsubsec:loco-manipulation}
Two priorities are considered in this work: manipulation and locomotion. 
\begin{itemize}
    \item \textit{Locomotion priority}: this mode is conceived for giving physical robotic assistance to the worker over large workspaces, e.g., carrying-and-lifting tasks, like moving a heavy object between places that are far away. In this mode, the mobile base's high-mobility capabilities are exploited to allow large movements over an ideally infinite workspace. The robot end-effector tracks the desired motion commanded by the human through the admittance interface while keeping a whole-body configuration that is comfortable for the users and does not hinder their natural movements. This configuration is called~\textit{preferred configuration} since it depends on human preferences and is affected by user-specific factors like favorite hand and person height. 
    
    \item \textit{Manipulation priority}: this mode is devised for performing conjoined manipulation actions and for changing the preferred robot configuration before switching to locomotion. Hence, the desired configuration is the current robot configuration when the switch to locomotion occurs. This feature presents the benefit of allowing each user to flexibly set the desired configuration before starting the locomotion depending on their preferences and on the specific requirements of the locomotion task. Note that, unexpected movements of the floating base are undesired since they might compromise manipulation accuracy.
\end{itemize}

\subsection{Supernumerary Bodies}
\label{subsec:supernumerary_bodies}
\begin{figure}
    \centering
    \includegraphics[width=0.8\columnwidth]{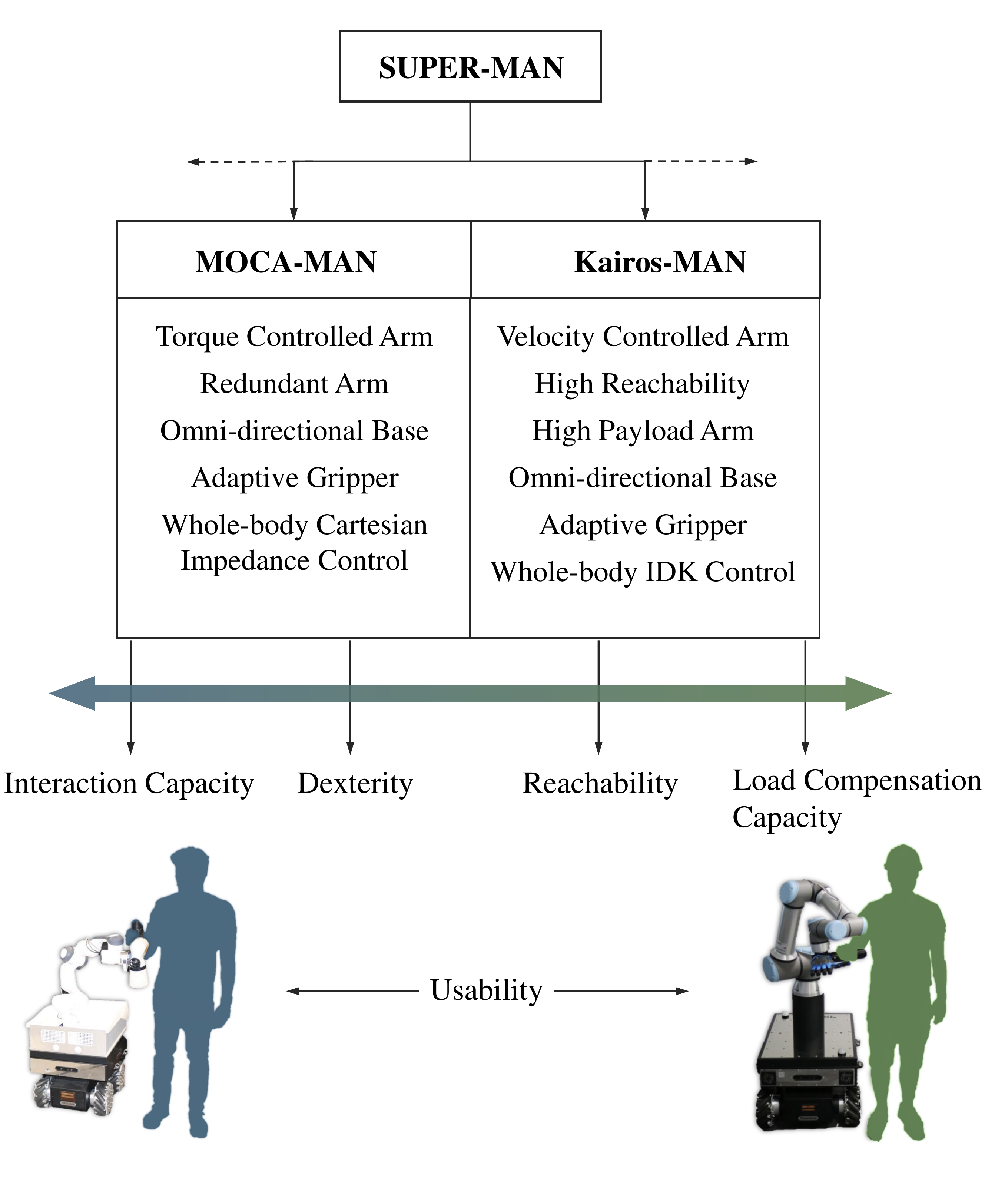}
    \caption{Illustration of similarities, differences and general features of the two particular supernumerary bodies considered in this work for experimentation. While MOCA-MAN (left-hand side) is more desirable for interactive tasks or task with high manipulability requirements, Kairos-MAN (right-hand side) is preferable for high-payload operations or those requiring higher reachability.}
    \label{fig:mocaman_vs_kairosman}
\end{figure}

The last element of the SUPER-MAN framework comprises a supernumerary robotic platform with loco-manipulation capabilities to provide physical support to the user. Many kinds of platforms can be integrated within the framework as explained in Section~\ref{sec:intro} and illustrated in Fig.~\ref{fig:mainfigure}. This work considers two wheeled-based mobile manipulators with different features: MOCA-MAN and Kairos-MAN. These two platforms and their characteristics are described below, and their differences and similarities are illustrated in Fig.~\ref{fig:mocaman_vs_kairosman}.


\begin{figure*}
    \centering
    \includegraphics[width=0.85\textwidth]{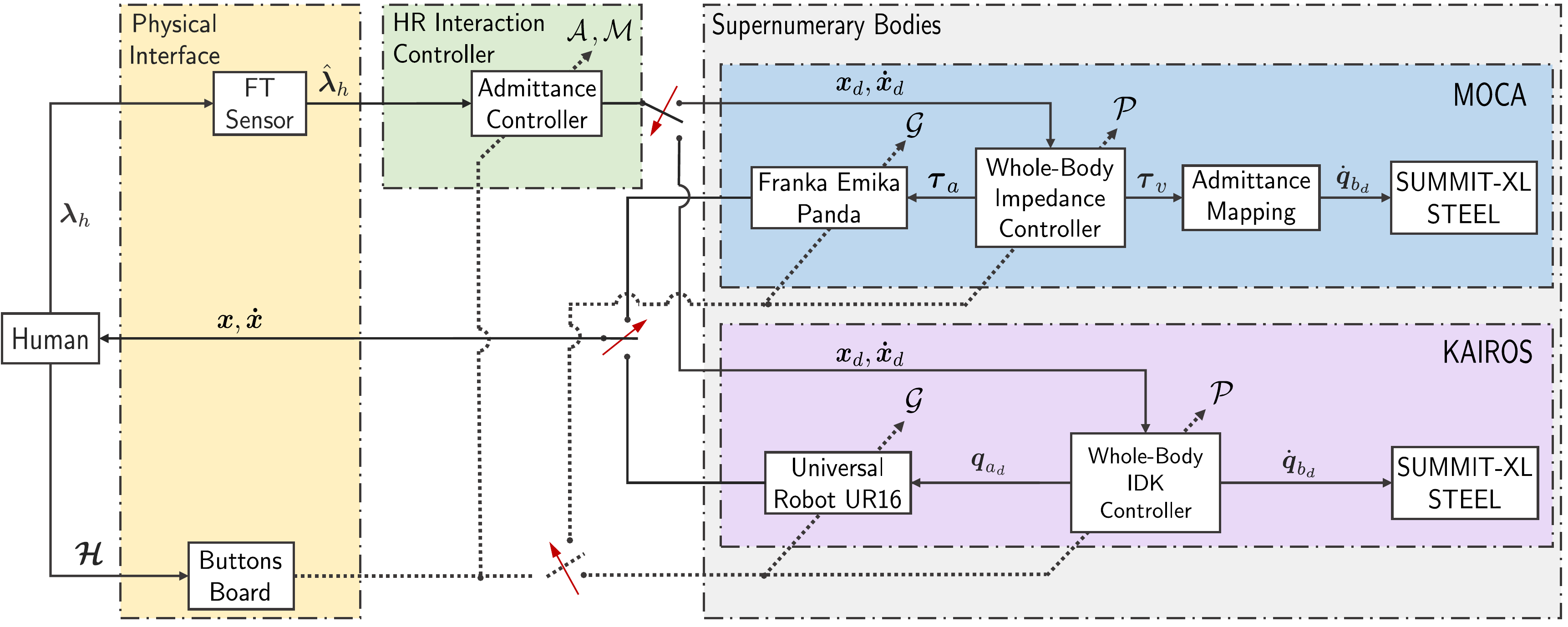}
    \caption{Block diagram of the SUPER-MAN framework. The red arrows indicate switches that represent the possibility of using the two robotic platforms and how the elements of the diagram interconnections and controllers change when using one robotic platform or the other (Switch up for MOCA and down for Kairos).}
    \label{fig:block_diagrams}
\end{figure*}

\subsubsection{The MOCA Robotic Platform}
\label{subsec:moca_robotic_platform}

\paragraph{Hardware Specifications:\\}
\label{par:hardware_moca}
The MOCA mobile manipulator (see Fig.~\ref{fig:systems}b) has been previously presented in~\cite{wu2019teleoperation}. This robot comprises a Robotnik SUMMIT-XL STEEL mobile platform and integrates a 7 DoFs Franka Emika Panda manipulator. The arm also has the Pisa/IIT SoftHand as a tool. The external wrenches are estimated from the torque sensors incorporated in the arm's joints, while the admittance interface is placed on the robotic arm's end-effector near the SoftHand. The complete platform is controlled by a whole-body Cartesian impedance controller described in detail in paragraph~\ref{par:controller_moca}. 

The reader can find the main characteristics of this platform listed in the left column at Fig.~\ref{fig:mocaman_vs_kairosman}. The main advantage of MOCA is the redundancy of the manipulator and the impedance controller, which ensures a compliant behavior of the manipulator when physical interactions with the environment occur. This feature is critical both for safety reasons and for executing conjoined human-robot manipulation tasks. Since the arm is redundant, during manipulation tasks, each configuration in the task space has more than one solution in the joint space. This characteristic increases the adaptability and comfort of the human partner. On the other hand, the robotic arm has a small payload of 3kg, limiting the number of industrial tasks one can carry out with the assistance of this platform.

\paragraph{MOCA Weighted Whole-body Cartesian Impedance Controller:\\}
\label{par:controller_moca}
Here, the weighted whole-body Cartesian impedance controller used on the MOCA robotic platform is described. A complete analysis of the controller can be found in our previous work~\cite{wu2021unified}. Fig.~\ref{fig:block_diagrams} depicts a schematic of the whole control framework, including the admittance-type interface and the HR interaction controller, where the blue rectangle includes the MOCA specific part.

In order to write the dynamics of the robot as a torque-controlled floating base system, an admittance control law is used to map virtual control torques into desired velocities of the velocity-controlled mobile base
\begin{equation}
\label{eq:admittance_eq_base}
    \boldsymbol{M}_{v}\boldsymbol{\ddot{q}}_{b} + \boldsymbol{D}_{v}\boldsymbol{\dot{q}}_{b} = \boldsymbol{\tau}_{v}.
\end{equation}
Then, the dynamics of the whole system can be written as
\begin{equation}
\begin{aligned}
\label{eq:whole_body_dynamics}
&  \begin{bmatrix}
 \boldsymbol{M}_v & \boldsymbol{0} \\
 \boldsymbol{0} & \boldsymbol{M}_a(\boldsymbol{q}_a)
\end{bmatrix} \boldsymbol{\ddot{q}} +
\begin{bmatrix}
 \boldsymbol{D}_v & \boldsymbol{0} \\
 \boldsymbol{0} & \boldsymbol{C}_a(\boldsymbol{q}_a,\dot{\boldsymbol{q}}_a)
\end{bmatrix} \boldsymbol{\dot{q}} + \\
& \begin{bmatrix}
 \boldsymbol{0} \\
 \boldsymbol{g}_a(\boldsymbol{q}_a)
\end{bmatrix} = \boldsymbol{\tau}_c + \boldsymbol{\tau}_{ext},
\end{aligned}
\end{equation}
with $\boldsymbol{\tau}_c = \begin{bmatrix} \boldsymbol{\tau}_v^T & \boldsymbol{\tau}_a^T \end{bmatrix}^T$ and $\boldsymbol{\tau}_{ext} = \begin{bmatrix} \boldsymbol{0}^T & \boldsymbol{\tau}_{a,ext}^T \end{bmatrix}^T$.

The whole-body Cartesian impedance controller generates high level torque references $\boldsymbol{\tau}_c$ that are then passed to the mobile base admittance controller (${\boldsymbol{\tau}_{v}}$)
and to the arm low-level controller (${\boldsymbol{\tau}_a}$). 
Such torques are defined as
(dependencies are dropped for sake of readability)
\begin{equation}\label{eq:opt_solution}
\begin{aligned}
    \boldsymbol{\tau}_c & = \boldsymbol{W^{-1}M^{-1}J^{T}\Lambda_{W}\Lambda^{-1}F} \\
    & + (\boldsymbol{I-W^{-1}M^{-1}J^{T}\Lambda_{W}JM^{-1}})\boldsymbol{\tau}_0 ,
\end{aligned}
\end{equation}
that fulfills the relationship between the generalized joint torques $\boldsymbol{\tau}_c$ and the generalized Cartesian forces $\boldsymbol{F}$ representing the desired impedance behavior $\boldsymbol{\bar{J}}^T \boldsymbol{\tau}_c = \boldsymbol{F} $, where
\begin{align}
    \notag \bar{\boldsymbol{J}} &= \boldsymbol{M}^{-1} \boldsymbol{J}^{T} \boldsymbol{\Lambda}  \; ,\\
    \boldsymbol{\Lambda_{W}} &= \boldsymbol{J}^{-T}\boldsymbol{MWM}\boldsymbol{J}^{-1}  \; ,\\
    \notag \boldsymbol{\Lambda} &= {\big( \boldsymbol{J}\boldsymbol{M}^{-1} \boldsymbol{J}^{T} \big)}^{-1}  \; .
\end{align}
The null-space torque $\boldsymbol{\tau}_0$ can be used to generate actions that do not interfere with the Cartesian force $\boldsymbol{F}$, since they are projected onto the null-space of the Cartesian task space.

The positive definite weighting matrix $\boldsymbol{W}$ is defined as
\begin{equation} \label{eq:weight}  
    \boldsymbol{W}(\boldsymbol{q})=\boldsymbol{H}^T\boldsymbol{M}^{-1}(\boldsymbol{q})\boldsymbol{H},
\end{equation}
where $\boldsymbol{H}$ is diagonal and dynamically selected according to the task:
\begin{equation} \label{eq:eta}
    \boldsymbol{H} = \begin{bmatrix} \eta_{B}\boldsymbol{I}_{n_b} & \boldsymbol{0}_{n_b\times n_a} \\ \boldsymbol{0}_{n_a\times n_b} & \eta_{ A } \boldsymbol{I}_{n_a} \end{bmatrix},
\end{equation}
where $\eta_{B}$ and $\eta_{A}$ penalize more or less the motion of base or arm, respectively.

Next, the desired Cartesian impedance behavior is obtained by
\begin{equation}
\boldsymbol{F} =   \boldsymbol{D}_d(\dot{\boldsymbol{x}}_d-\dot{\boldsymbol{x}}) + \boldsymbol{K}_d(\boldsymbol{x}_d - \boldsymbol{x}).
    \label{eq:cartesian_impedance}
\end{equation}
Finally, the null-space torque $\boldsymbol{\tau}_0$ is generated as
\begin{equation} \label{eq:nullspace_impedance}
  \boldsymbol{\tau}_{0} = -\boldsymbol{D}_0\dot{\boldsymbol{q}} + \boldsymbol{K}_0(\boldsymbol{q}_{pref} - \boldsymbol{q}).
\end{equation}
The choice of $\boldsymbol{K}_0$ is important for the determination of the loco-manipulation behavior of the robot. In particular, high coefficients in $\boldsymbol{K}_0$ let the base move in order to keep $\boldsymbol{q}$ close to $\boldsymbol{q}_{pref}$, allowing a predominant locomotion behavior, vice-versa if the coefficients in $\boldsymbol{K}_0$ are small.

\subsubsection{The Kairos Robotic Platform}
\label{subsec:Kairos}

\paragraph{Hardware Specifications:\\}
\label{par:hardware_kairos}
The Kairos mobile manipulator (Fig.~\ref{fig:systems}c) comprises a Robotnik SUMMIT-XL STEEL mobile platform and a 6 DoFs Universal Robot UR16e manipulator. This manipulator also integrates a F/T sensor that measures the applied wrenches at the robot's flange. Hence, the manipulator integrates two F/T sensors, one in the admittance interface to measure human forces, and one at the end-effector to measure external forces due to interactions with the environment. The complete system is controlled by a whole-body controller described in detail in paragraph~\ref{par:controller_kairos}. The robotic arm also integrates the Pisa/IIT SoftHand for grasping purposes. The admittance interface is placed on the robotic arm's end-effector near the SoftHand.  

The main features of this robotic platform are listed in the right column at Fig.~\ref{fig:mocaman_vs_kairosman}. The principal benefit of using this platform is the high payload (16kg) of the robotic arm that allows the user to carry heavy objects or handle heavy tools without significant physical effort. On the other hand, the robotic arm is a non-redundant 6 DoFs robotic manipulator. Hence, during manipulation, each configuration in the task space has only one solution in the joint space, resulting in a loss of dexterity and discomfort on the human partner.

\paragraph{Kairos Weighted Whole-body Closed-Loop Inverse Differential Kinematics Controller:\\}
\label{par:controller_kairos}

Here, the weighted whole-body closed-loop inverse differential kinematics controller used on the Kairos robotic platform is described. In Fig.~\ref{fig:block_diagrams}, the Kairos specific part is depicted in purple.


\begin{table*}
    \caption{Results of the Prolonged Manipulation Experiment}
    \label{tab:resultsMOCA}
    \centering
    \begin{tabular}{c|c|c|c|c|c|c|c|c|c|c|c|c|c|c}
     & S1  & S2 & S3 & S4 & S5 & S6  & S7 & S8 & S9 & S10 & S11 & S12 & Mean (stdev) & SS\\
    \hline \hline 
    
    & 0 & 0 & 0 & 0 & 1 & 0 & 1 & 1 & 2 & 2 & 0 & 0 & 0.58 (0.79) \\ \rowcolor[HTML]{EFEFEF}
    \cellcolor{white}\multirow{-2}{*}{$N_{err}$  } & 2 & 1 & 0 & 2 & 4 & 0 & 1 & 0 & 4 & 0 & 0 & 1 & 1.25 (1.48) & \cellcolor{white}\multirow{-2}{*}{No}\\ \hline
    
    & 177 & 207 & 143 & 184 & 177 & 173 & 174 & 148 & 223 & 209 & 178 & 176 & 180.75 (23.25) \\ \rowcolor[HTML]{EFEFEF}
    \cellcolor{white}\multirow{-2}{*}{$T$ [s]} & 128 & 122 & 114 & 135 & 124 & 122 & 122 & 141 & 155 & 143 & 126 & 124 & 129.58 (11.64) & \cellcolor{white}\multirow{-2}{*}{Yes} \\ \hline

    \end{tabular}
    \begin{flushleft}
    \hspace{0.2cm}    $N_{err}$: Number of errors.
    \hspace{1.05cm}    $T$: Completion Time. 
    \hspace{3.05cm}    \textit{Background color code:} white--HR, gray--H\\
    \hspace{0.2cm}    S1 -- S12: Subjects.  
    \hspace{1.45cm} SS: Statistical Significance ($p<0.001$)\\
    \end{flushleft}
\end{table*}
%
The controller solves a weighted inverse differential kinematics problem while exploiting the redundancy of the robot to fulfill a secondary task. This problem can be formulated as a hierarchical quadratic program (HQP). The different requirements are expressed as quadratic cost functions to be optimized with different levels of priority: tasks with lower priority are possibly realized in the null-space of higher priority tasks. The more redundant DoFs the robot has, the richer the HQP formulation can be, e.g., there can be more than two priority levels, and equality and inequality constraints can be included~\cite{Tassi2021}. 
The whole-body controller computes the following solution for the robot joints velocities~\cite{kazerounian1988global}
\begin{equation}
    \boldsymbol{\dot{q}}_d= \boldsymbol{\dot{q}}_{d,1} + \mathcal{N}(\boldsymbol{J}(\boldsymbol{q}))\boldsymbol{\dot{q}}_{d,2},
\end{equation}
where $\boldsymbol{\dot{q}}_{d,1}$ and $\boldsymbol{\dot{q}}_{d,2}$ are the minimizers of the cost functions representing the primary and the secondary tasks, respectively. The primary cost function is written as
\begin{equation}
\label{eq:firsttaskargmin}
    \mathcal{L}_1(\boldsymbol{\dot{q}}) = \frac{1}{2}[\norm{\boldsymbol{\dot{x}}_d+\boldsymbol{K}(\boldsymbol{x}_d-\boldsymbol{x})-\boldsymbol{J}(\boldsymbol{q})\boldsymbol{\dot{q}}}_{\boldsymbol{W}_1}^2 +\norm{\boldsymbol{\dot{q}}}_{\boldsymbol{W}_2}^2],
\end{equation}
where $\boldsymbol{W}_2 = diag\{w_b \boldsymbol{1}_{n_b},w_ak^2\boldsymbol{1}_{n_a}\}$ with $k$ being the damping factor~\cite{wampler1986manipulator, chiaverini1992weighted}. The first quadratic cost in Equation~\ref{eq:firsttaskargmin} is responsible for solving the Closed Loop Inverse Kinematics (CLIK) problem~\cite{chiacchio1991closed}, where $\boldsymbol{W}_1$ weights the relative tracking importance of the 6 DoFs. Instead, the second quadratic cost is a regularization term used for improving numerical stability, for guaranteeing a trade-off between tracking accuracy and solution feasibility through online tuning of $k$~\cite{chiaverini1997singularity} and for exploiting more or less specific joints through $\boldsymbol{W}_2$, e.g., the ones of the mobile base ($w_b<w_a$) or of the arm ($w_a<w_b$). In particular, $k$ is tuned online as a function of the manipulability index of the arm $w(\boldsymbol{q}_a)$~\cite{deo1995overview, wampler1986manipulator, hollerbach1987redundancy}
\begin{equation}
    w(\boldsymbol{q}_a)=\sqrt{\det(\boldsymbol{J}_{a}(\boldsymbol{q}_a)\boldsymbol{J}_{a}(\boldsymbol{q}_a)^T)}.
\end{equation}
To avoid unfeasible solutions (infinite joint velocities) for tracking the desired motion at the end-effector, $k$ is tuned online so that the relative importance of minimizing the joint velocities with respect to accomplishing tracking of the desired end-effector motion is increased when close to singularity~\cite{deo1995overview, wampler1986manipulator, seraji1990improved}:
\begin{equation}
    k = \begin{cases} 1+k_0(1-\frac{w(\boldsymbol{q}_a)}{w_t})^2 & \mbox{if } w(\boldsymbol{q}_a) \le w_t \\ 1 & \mbox{otherwise. } \end{cases}
\end{equation}
Note that $k$ penalizes the arm joints and it is tuned considering only the arm configuration since the base does not feature any singularity. Hence, the minimizer of equation~\ref{eq:firsttaskargmin} is~\cite{seraji1990improved}
\begin{equation}
\label{eq:firsttasksolution}
\begin{aligned}
     \boldsymbol{\dot{q}}_{d,1} = & (\boldsymbol{J}(\boldsymbol{q})^T\boldsymbol{W}_1\boldsymbol{J}(\boldsymbol{q})+\boldsymbol{W}_2)^{-1} \\ 
    & \boldsymbol{J}(\boldsymbol{q})^T\boldsymbol{W}_1[\boldsymbol{\dot{x}}_d+\boldsymbol{K}(\boldsymbol{x}_d-\boldsymbol{x})]. \\
\end{aligned}
\end{equation}
Next, $\boldsymbol{\dot{q}}_{d,2}$ is computed in order to minimize the secondary cost function~\cite{nakanishi2005comparative}
\begin{equation}
\label{eq:secondarycostfunctionlocom}
    \mathcal{L}_{2}(\boldsymbol{q}) = \frac{1}{2}\norm{\boldsymbol{q}_{pref}-\boldsymbol{q}}_{\boldsymbol{G}_i}^2,
\end{equation}
where $\boldsymbol{G}_i=diag\{\boldsymbol{0}_{n_b},k_i\boldsymbol{1}_{n_a}\}$. The first $n_b$ terms of $\boldsymbol{G}_i$ are always zero since a preferred configuration of the floating base never exists. Indeed, the mobile base should not try to stay close to a certain fixed position and orientation in the world. Rather, the arm should remain close to a certain configuration relative to the base. Indeed, the last $n_a$ terms are equal to $k_i$, that varies according to the priority mode selected. In particular:
\begin{equation}
    k_i \begin{cases} > 0 & \mbox{if locomotion}  \\ = 0 & \mbox{if manipulation.} \end{cases}
\end{equation}
Therefore, during manipulation, zero weight is assigned to the secondary task since there is no preferred configuration.
Hence, $\boldsymbol{\dot{q}}_{d,2}$ is found as~\cite{chiaverini1997singularity}
\begin{equation}
    \boldsymbol{\dot{q}}_{d,2} = - (\frac{d\mathcal{L}_2(\boldsymbol{q})}{d\boldsymbol{q}})^T= \boldsymbol{G}_i(\boldsymbol{q}_{pref}-\boldsymbol{q}).
\end{equation}
%

\section{Experiments and Results}
\label{sec:experiments}
In this section, four experimental studies and their results are delivered. Two user studies were carried out to evaluate the performance and user acceptance of the two supernumerary systems. Tasks matching the properties of the corresponding robotic platform were selected. Particularly, for MOCA-MAN a prolonged manipulation task was considered that needs human-level dexterity in lengthy and medium/low effort tasks. While for Kairos-MAN, an overhead drilling operation was selected that is heavily effort demanding and places an extremely high level of stress on the human upper body. This kind of task is widely used in Human-Robot Interaction research~\cite{guler2022adaptive}. The choices of the whole-body controller and admittance controller parameters are reported in Appendix~\ref{appendix:controller_params}. Two additional experiments are reported to reveal the potential of the SUPER-MAN approach during the execution of constrained interactive and long distance load-carrying tasks. A video of the experiments is available in our youtube channel~\footnote{The video can be found at \url{https://youtu.be/_kfhLYQjhvA}}.

\subsection{Usability Study}
\label{subsec:User_Study}

Twelve healthy volunteers, six males and six females, (age: $27.8 \pm 1.8$ years; mass: $65.2 \pm 16.2$ kg; height: $172.0 \pm 10.1$ cm)\footnote{Subject data is reported as: mean $\pm$ standard deviation.} were recruited for the user studies. 
Participants were students and researchers with no or limited experience in industrial work. 
After explaining the experimental procedure, written informed consent was obtained, and a numerical ID was assigned to anonymize the data. The whole experimental activity was carried out at the Human-Robot Interfaces and Interaction (HRII) Lab, Istituto Italiano di Tecnologia (IIT), in accordance with the Declaration of Helsinki. The protocol was approved by the ethics committee Azienda Sanitaria Locale (ASL) Genovese N.3 (Protocol IIT\_HRII\_ERGOLEAN 156/2020).

\begin{figure}
    \centering
    \includegraphics[width=1\columnwidth]{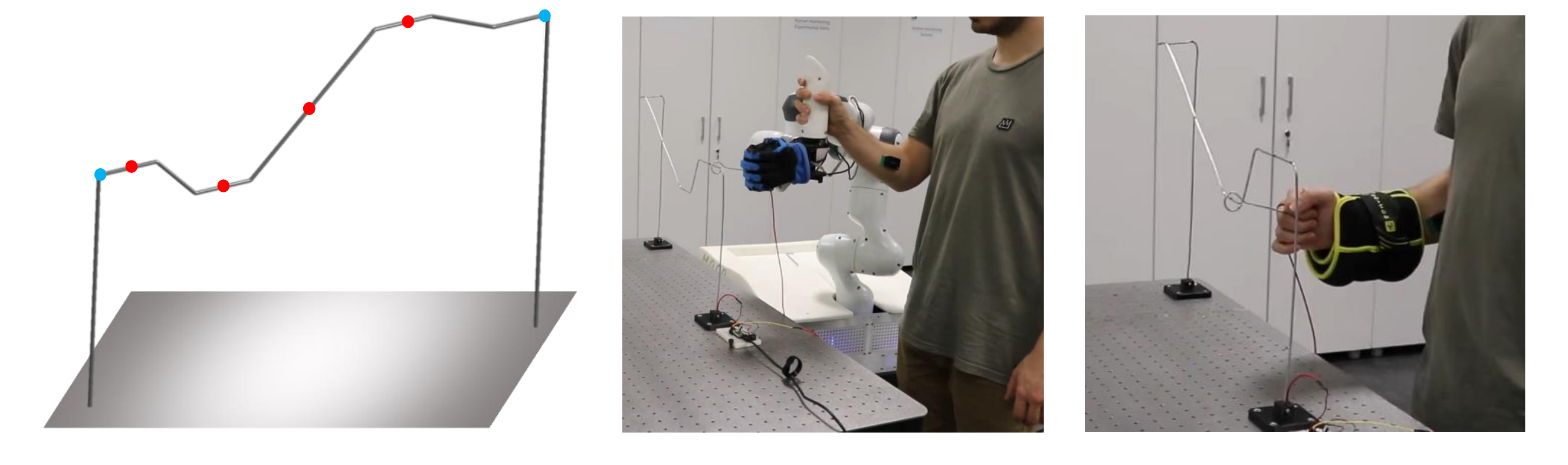}\\
    \hspace{0.5cm} (a) \hspace{2.3cm} (b) \hspace{2.1cm} (c)
    \caption{Experimental setup of the MOCA-MAN user study (Prolonged Manipulation): (a) A CAD file of the track. The blue spots represent the initial (left) and turning (right) points of the path, whereas the red ones represent the pause points. (b) A user during the experiment with robot assistance. (c) A user during the experiment without robot assistance. An additional load of 3kg is attached to the user's hand to simulate a tool weight.}
    \label{fig:expSetupMOCAMAN}
\end{figure}

\subsubsection{Prolonged Manipulation}
A 3D path tracking task was designed to evaluate the potential of the SUPER-MAN framework in performing tasks that require prolonged and precise arm movements while using medium/low-weight tools (e.g., painting, polishing, etc.). In this experiment, the MOCA-MAN setting was chosen, due to the arm redundancy and better interaction capacity of the torque-controlled arm, while the human-robot (HR) conjoined case was compared against the human alone (H) case as shown in Fig.~\ref{fig:expSetupMOCAMAN}. In particular, the path to be tracked (Fig.~\ref{fig:expSetupMOCAMAN}a), and the setup in HR (Fig.~\ref{fig:expSetupMOCAMAN}b) and H (Fig.~\ref{fig:expSetupMOCAMAN}c) are shown. In HR, the robot carried the tool while the human guided it using the admittance interface. Since the robot can sustain a 3kg tool, in H, the human wrist was loaded with a 3kg mass to simulate the weight of a tool. Please note that in HR this additional load is not considered as it would be carried by the robot arm, and for this particular case the influence of the weight is negligible.
A ring-shaped tool of $~3cm$ diameter was used to have a certain tolerance in the tracking precision while easing the detection of an error. When the tool touched the wire, a sound warned the participants (similar to the buzz wire game). Referring to Fig.~\ref{fig:expSetupMOCAMAN}a, the subjects were asked to track the path going back and forth one time in between the two blue dots. Besides, they were asked to stop for 10 seconds every time a red dot was crossed. 
The subjects were informed that their performance was evaluated on completion time and number of errors. Before each trial of the experiment (H and HR), the participants were given some time for training to avoid learning effects and to familiarize themselves with the platform. The training time varies among subjects, as the learning ability of each subject is subjective. Therefore, none of the subjects was forced to perform the experiment without feeling sufficiently familiarized, nor did we want to push anyone to perform more training time, despite feeling comfortable already. Moreover, participants could rest between experimental trials to avoid undesired fatigue effects.

The completion time ($T$) and the number of errors ($N_{err}$) were used as performance metrics. In addition, for each condition, the subjects filled in the NASA-TLX questionnaire, where the users rated the workload from 0 to 100 for six subscales representing different workload shades. The muscular activities of Anterior Deltoid (AD), Posterior Deltoid (PD), Biceps (BC), and Triceps (TR) were recorded using the Delsys Trigno platform, a wireless sEMG system commercialized by Delsys Inc. (Natick, MA, United States). Next, the signals obtained were filtered and normalized to their Maximum Voluntary Contractions (MVC). 

\begin{figure}
    \centering
    \includegraphics[width=0.8\columnwidth]{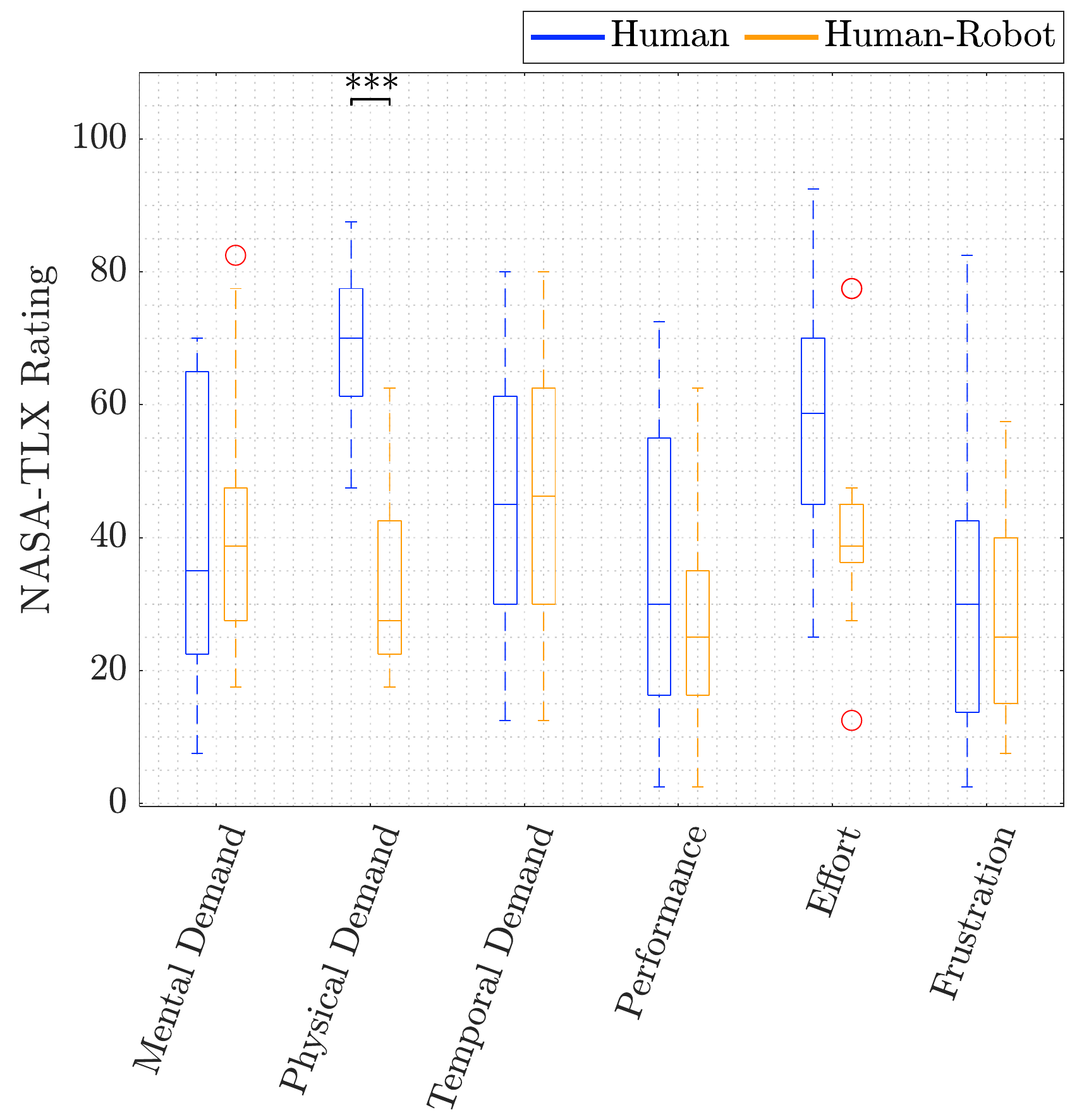}
    \caption{MOCA-MAN user study (Prolonged Manipulation): NASA-TLX questionnaire outcomes for H and HR. The statistical significance of the results obtained for each perceived workload index is tested using a sign test. *:$p<0.05$, **:$p<0.01$, ***:$p<0.001$, nothing: not significant.}
    \label{fig:NASATLX_MOCAMAN}
\end{figure}

\begin{figure}[t]
    \centering
    \includegraphics[width=0.8\columnwidth]{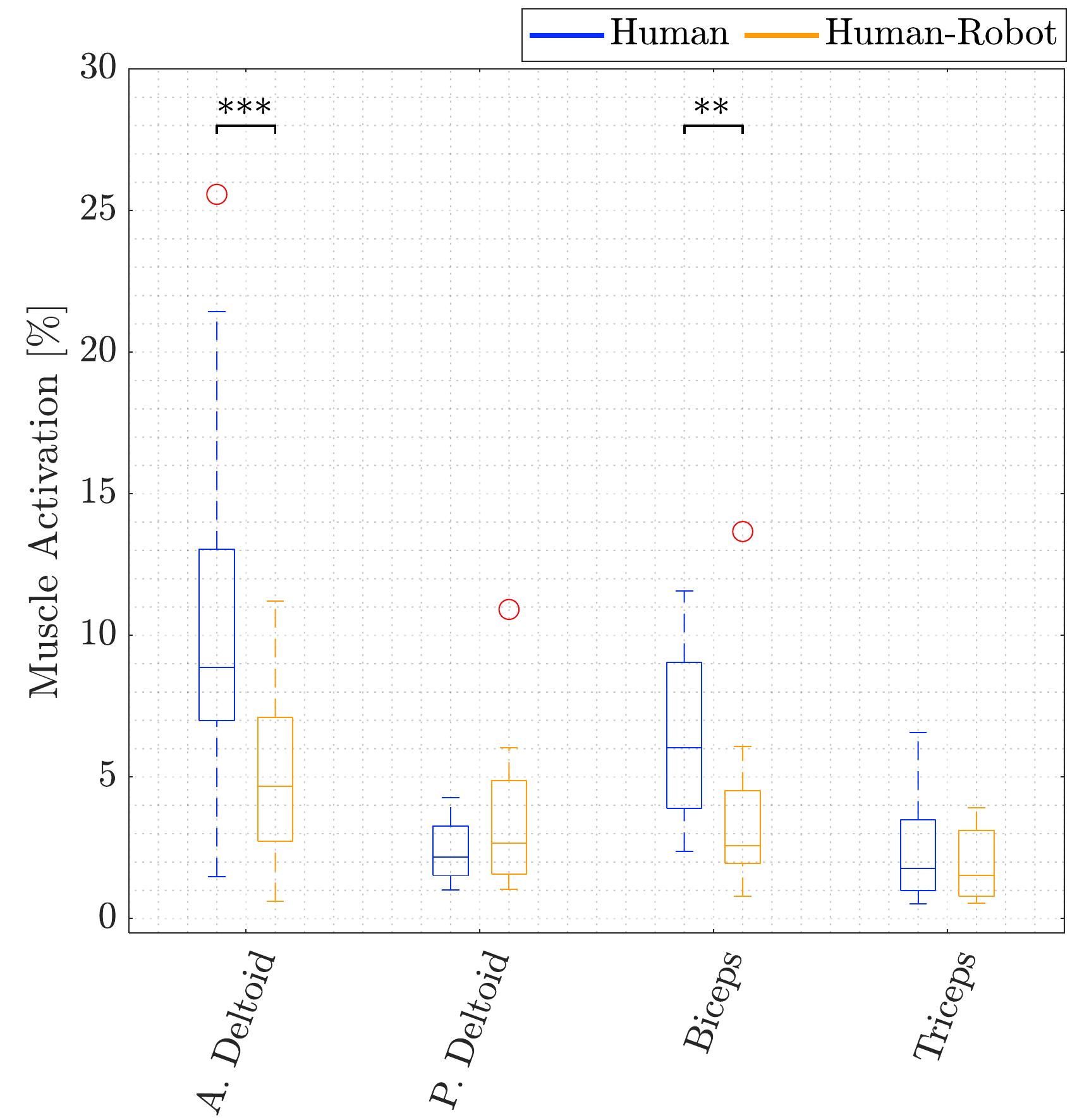}
    \caption{MOCA-MAN user study (Prolonged Manipulation): sEMG average values over the whole task for four muscles (AD, PD, BC and TR) for H and HR. The statistical significance of the results obtained for each muscle is tested using a sign test. *:$p<0.05$, **:$p<0.01$, ***:$p<0.001$, nothing: not significant.}
    \label{fig:EMGs_MOCAMAN}
\end{figure}

\begin{table*}
    \caption{Results of the Overhead Drilling Experiment}
    \label{tab:resultsKairos}
    \centering
    \begin{tabular}{c|c|c|c|c|c|c|c|c|c|c|c|c|c|c}
     & S1  & S2 & S3 & S4 & S5 & S6  & S7 & S8 & S9 & S10 & S11 & S12 & Mean (stdev) & SS \\
    \hline \hline 
    
    & 20 & 20 & 20 & 20 & 20 & 20 & 20 & 20 & 20 & 20 & 20 & 20 & 20 (0) \\ \rowcolor[HTML]{EFEFEF}
    \cellcolor{white}\multirow{-2}{*}{$N_{Hl}$} & 10 & 20 & 11 & 10 & 13 & 13 & 20 & 6 & 6 & 14 & 13 & 20 & 13 (4.93) & \cellcolor{white}\multirow{-2}{*}{Yes}\\ \hline
    
    & 12.93 & 9.20 & 10.75 & 9.46 & 9.67 & 10.68 & 9.90 & 11.00 & 9.70 & 8.75 & 9.10 & 9.30 & 10.04 (1.15) \\ \rowcolor[HTML]{EFEFEF}
    \cellcolor{white}\multirow{-2}{*}{$\overline{T}$ [s]} & 8.00 & 7.25 & 7.91 & 7.30 & 7.46 & 7.92 & 7.35 & 8.50 & 8.67 & 7.36 & 7.57 & 6.85 & 7.68 (0.53) & \cellcolor{white}\multirow{-2}{*}{Yes} \\ \hline
    
    & 0 & 1 & 0 & 0 & 0 & 0 & 1 & 1 & 1 & 1 & 0 & 0 & 0.42 (0.51) \\ \rowcolor[HTML]{EFEFEF}
    \cellcolor{white}\multirow{-2}{*}{$N_{err}$} & 2 & 7 & 3 & 0 & 0 & 1 & 4 & 0 & 0 & 1 & 0 & 8 & 2.17 (2.82) & \cellcolor{white}\multirow{-2}{*}{Yes} \\ 
    
    \end{tabular}

    \begin{flushleft}
    \hspace{0cm}    $N_{Hl}$: Number of holes completed.
    \hspace{0.5cm}    $\overline{T}$: Normalized completion time. 
    \hspace{0.5cm}    Background color code: white--HR, gray--H.\\
    \hspace{0cm}    $N_{err}$: Number of errors.
    \hspace{2cm}     S1 -- S12: Subjects.\\
    \hspace{0cm}  SS: Statistical Significance ($p<0.001$ in the case of $\overline{T}$, and  $p<0.01$ otherwise)
    \end{flushleft}
\end{table*}

\begin{figure}
    \centering
    \includegraphics[width=0.9\columnwidth]{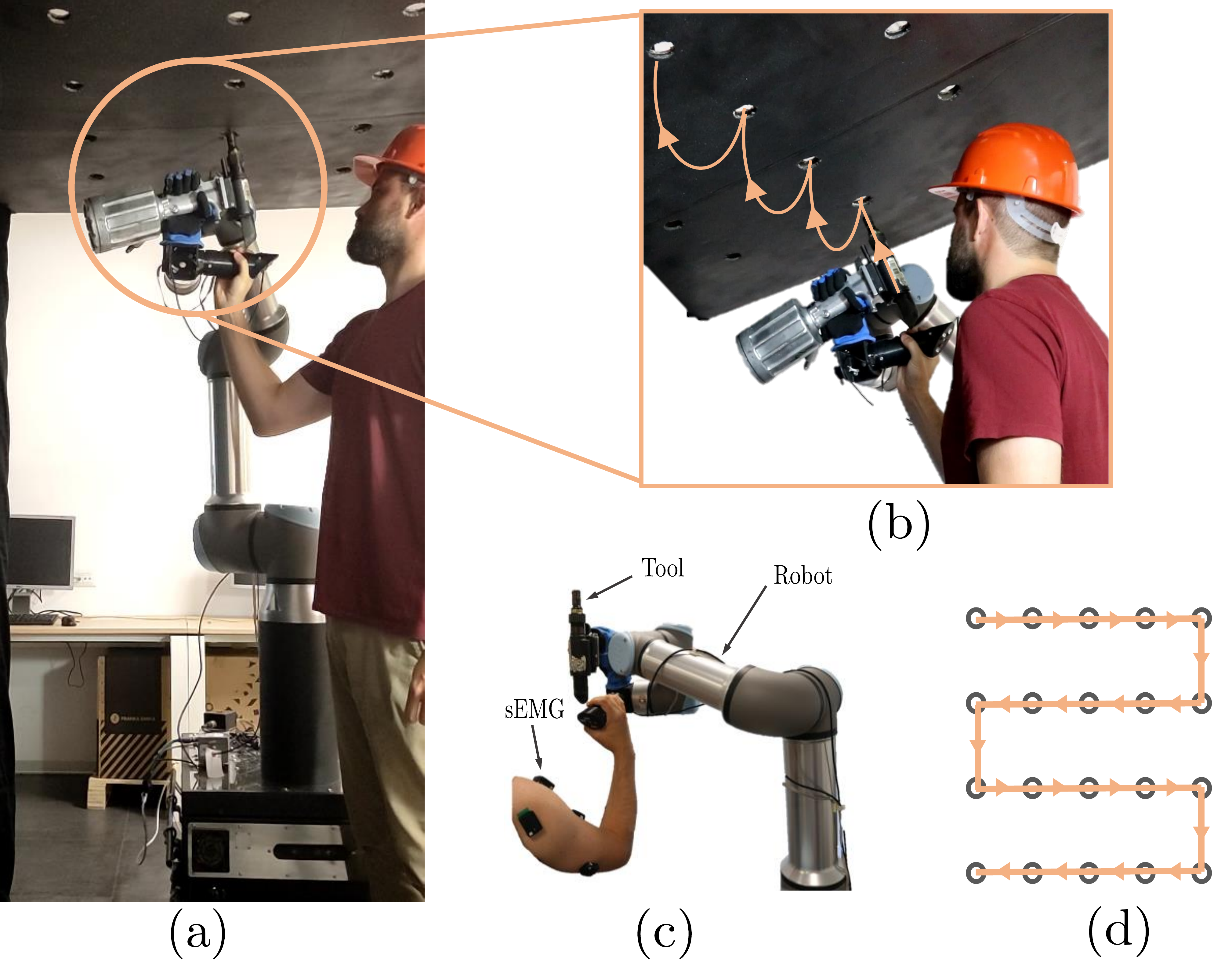}
    \caption{Experimental setup of the Overhead Drilling Task: (a) A worker during the experiment with robot assistance; (b) An zoomed picture of the experimental setup, (c) A zoomed picture of the sEMGs attached to the worker arm while performing the task with robot assistance; (c) A detailed sketch of the task path.}
    \label{fig:expSetup}
\end{figure}

The outcomes of the performance metrics for the 12 participants are reported in Table~\ref{tab:resultsMOCA} and the results of the NASA-TLX questionnaire and the sEMG average values during the task are shown in the boxplots of Fig.~\ref{fig:NASATLX_MOCAMAN} and~\ref{fig:EMGs_MOCAMAN}, respectively. The statistical significance of the results was tested using sign tests. The performance data show that the average number of errors was lower in HR, whereas the mean completion time was lower in H. However, statistical significance was found only for the completion time ($p<0.001$). Then, the average sEMG values for AD and BC were lower in HR with $p<0.001$ and $p<0.01$, respectively. No statistical significance was found for PD and TR. Regarding the NASA-TLX questionnaire, the Physical Demand was the only perceived workload dimension that showed a statistically significant difference ($p<0.001$), where HR obtained a better rating than H.

\subsubsection{Overhead Drilling}

The task consisted of a simulated overhead drilling operation, which is commonly used to evaluate the performance of the upper body exoskeletons. Similar to the previous user study, the cases of HR and H were compared, where in HR the human guided the robot through the admittance interface, and the robot carried a 5kg tool, while in H, the human performed the task alone. A plate having 20 holes of $7cm$ diameter, separated $30cm$ from each other, was hung above the participants' heads. The subjects had to insert the tool in each hole and keep it inside for 5 seconds without touching the plate. If the plate was touched, an error was counted. When the 20 holes were completed, or the subjects could not keep doing the task due to fatigue, the task was finished. The subjects were asked to minimize the completion time and the number of errors, and complete as many holes as possible. In Fig.~\ref{fig:expSetup} the experimental setup for HR and a graphical representation of the task are depicted. Before each trial of the experiment (H and HR), the participants were given some time for training to avoid learning effects and familiarize themselves with the platform. Moreover, participants could rest between experimental trials to avoid undesired fatigue effects. Due to the large reachability and payload requirements, the Kairos-MAN setup was used in this experiment.

As performance metrics, normalized completion time ($\overline{T}$), number of holes completed ($N_{Hl}$) and number of errors ($N_{err}$) were used. In order to have a fair comparison of the completion time, it was normalized over the number of holes completed. Muscular activity measurements were obtained in four arm locations, i.e., AD, PD, BC, and TR, through the same sEMG sensors used in the user study conducted on MOCA. Afterward, the signals were filtered and normalized to Maximum Voluntary Contractions (MVC). In addition, participants underwent the NASA-TLX questionnaire to evaluate users' subjective perception of workload in the two conditions considered. 

\begin{figure}
    \centering
    \includegraphics[width=0.8\columnwidth]{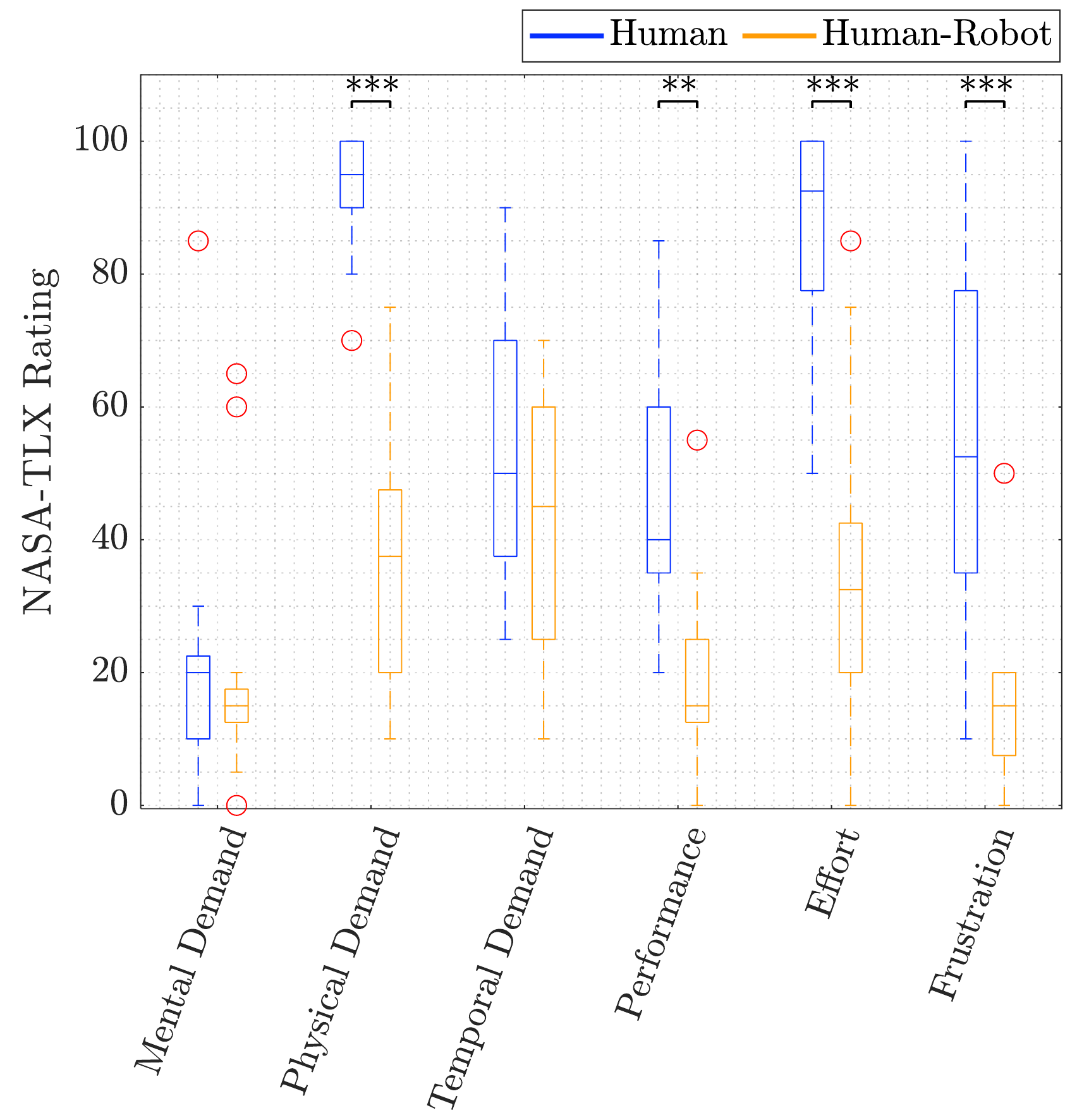}
    \caption{Kairos-MAN user study (Overhead Drilling): NASA-TLX questionnaire outcomes for H and HR. The statistical significance of the results obtained for each perceived workload index is tested using a sign test. *:$p<0.05$, **:$p<0.01$, ***:$p<0.001$, nothing: not significant.}
    \label{fig:NASATLX_KairosMAN}
\end{figure}

\begin{figure}[t]
    \centering
    \includegraphics[width=0.8\columnwidth]{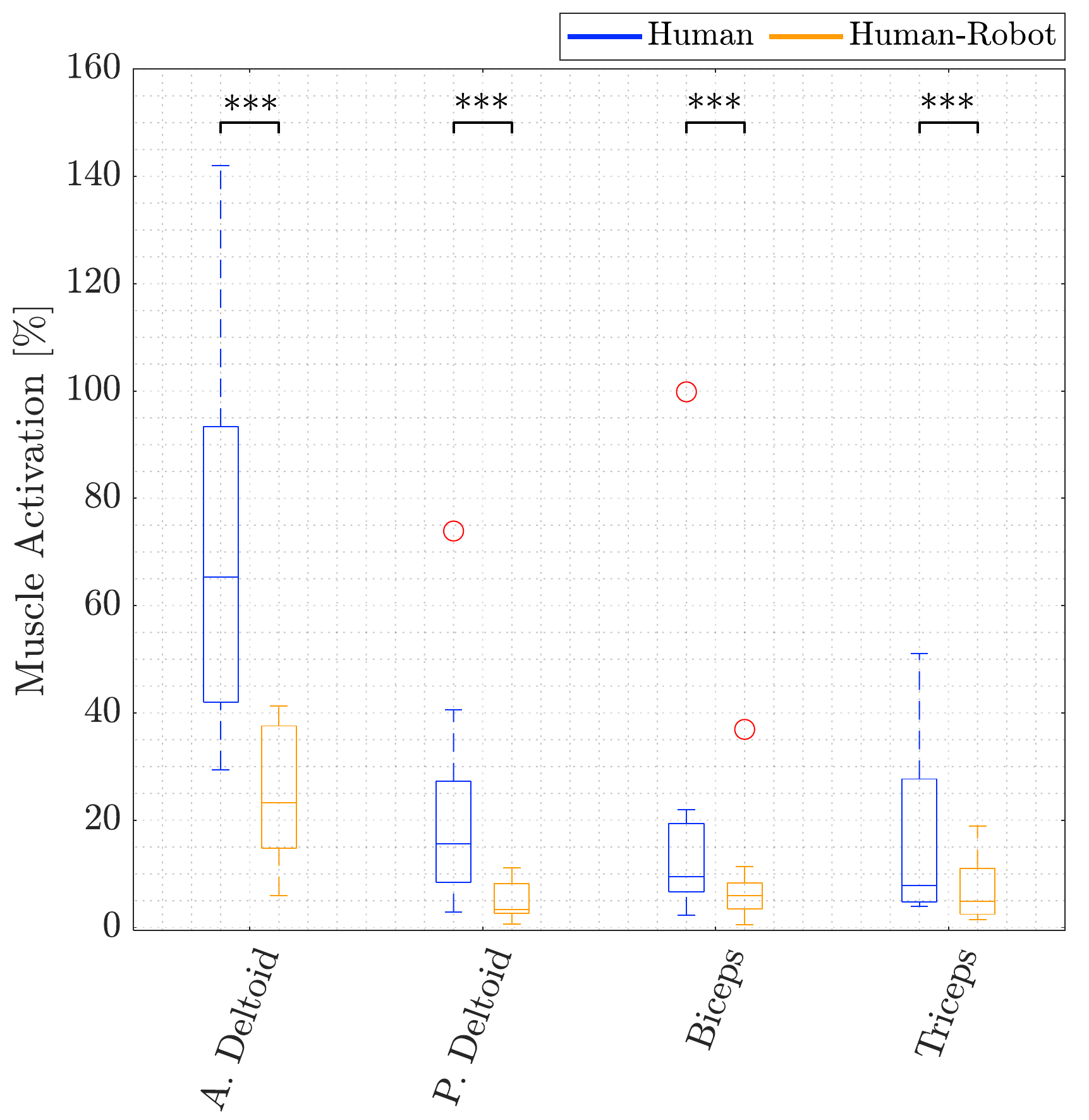}
    \caption{Kairos-MAN user study (Overhead Drilling): sEMG average values over the whole task for four muscles (AD, PD, BC and TR) for H and HR. The statistical significance of the results obtained for each muscle is tested using a sign test. *:$p<0.05$, **:$p<0.01$, ***:$p<0.001$, nothing: not significant.}
    \label{fig:EMGs_KairosMAN}
\end{figure}

Table~\ref{tab:resultsKairos} reports the performance metrics results. Fig.~\ref{fig:NASATLX_KairosMAN} and~\ref{fig:EMGs_KairosMAN} show the outcomes of the NASA-TLX questionnaire and average sEMG measurements through boxplots, respectively. The statistical significance of the results was tested using sign tests. The users perceived a higher workload in terms of Physical Demand ($p<0.001$), Performance ($p<0.01$), Effort ($p<0.001$) and Frustration ($p<0.001$) in H, while no statistical significance was found for perceived Mental Demand and Temporal Demand. Regarding the average sEMG measurements, HR featured a significant lower effort for all the muscles ($p<0.001$). Finally, HR outperformed H for $N_{err}$ and $N_{Hl}$ ($p<0.01$), while H presented smaller $\overline{T}$ ($p<0.001$) w.r.t. HR.

\begin{figure}
    \centering
    \includegraphics[width=1\columnwidth]{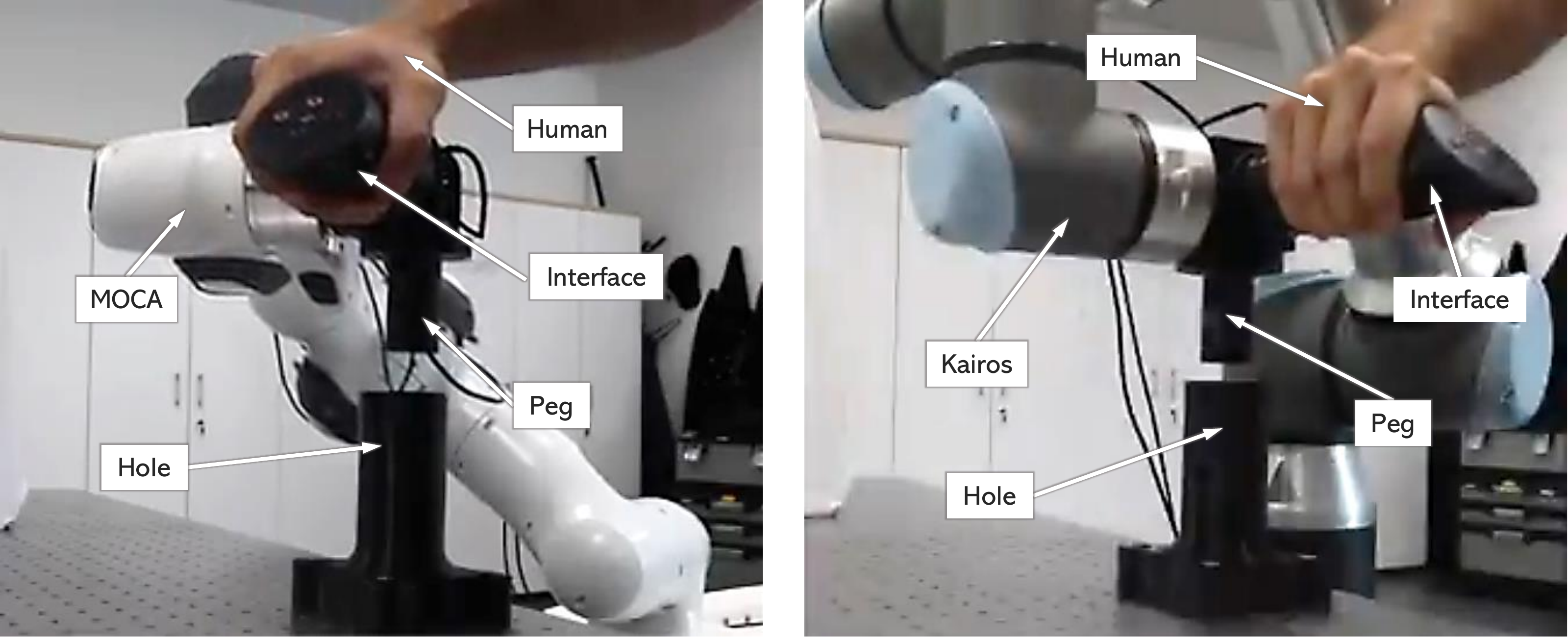}
    \caption{The experimental setup for Constrained Interaction trials. Pictures taken during the experiment with MOCA-MAN (left) and Kairos-MAN (right).}
    \label{fig:peginhole}
\end{figure}

\begin{figure*}
    \centering
    \includegraphics[width=0.9\columnwidth]{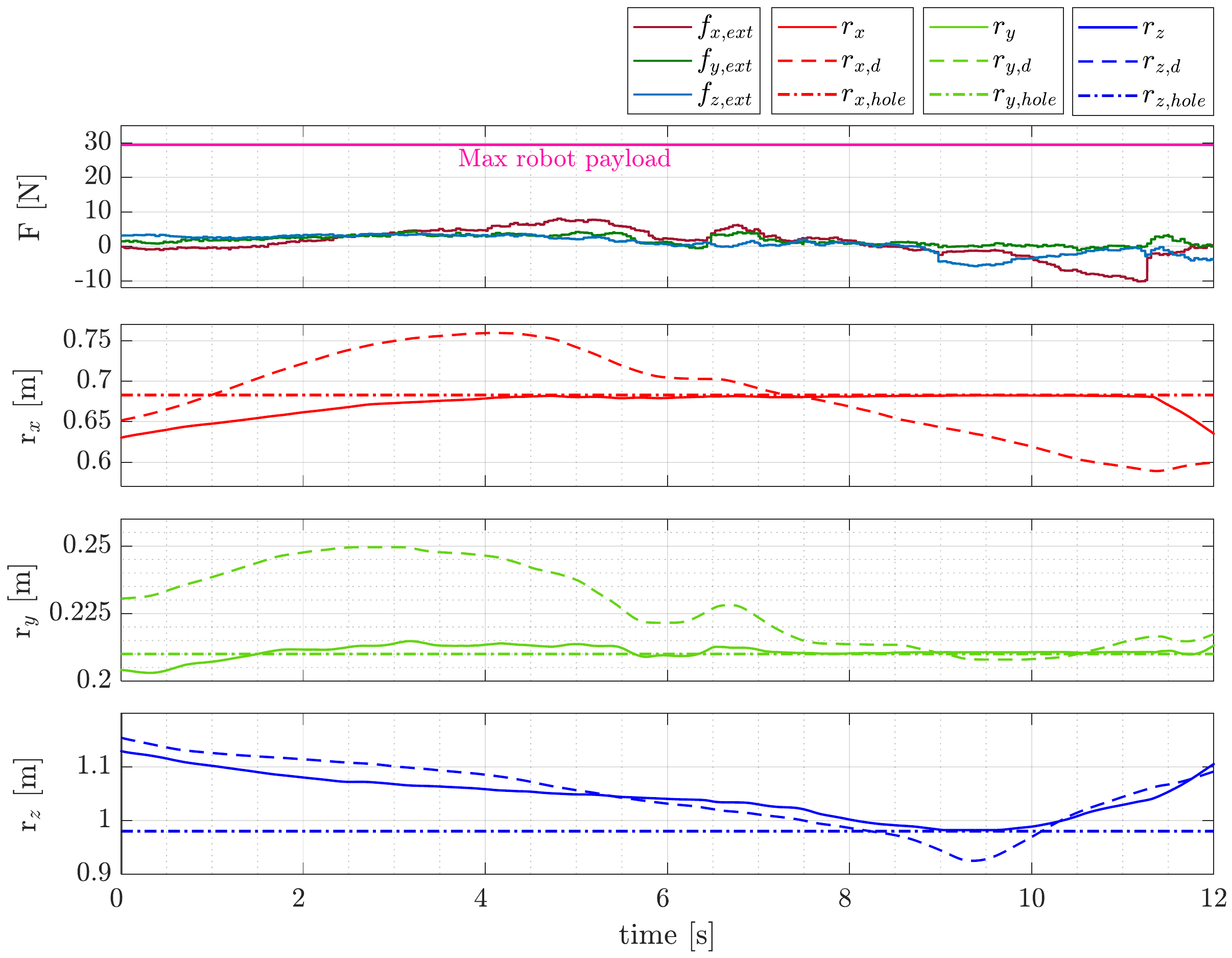}
    \hspace{0.7cm}
    \includegraphics[width=0.9\columnwidth]{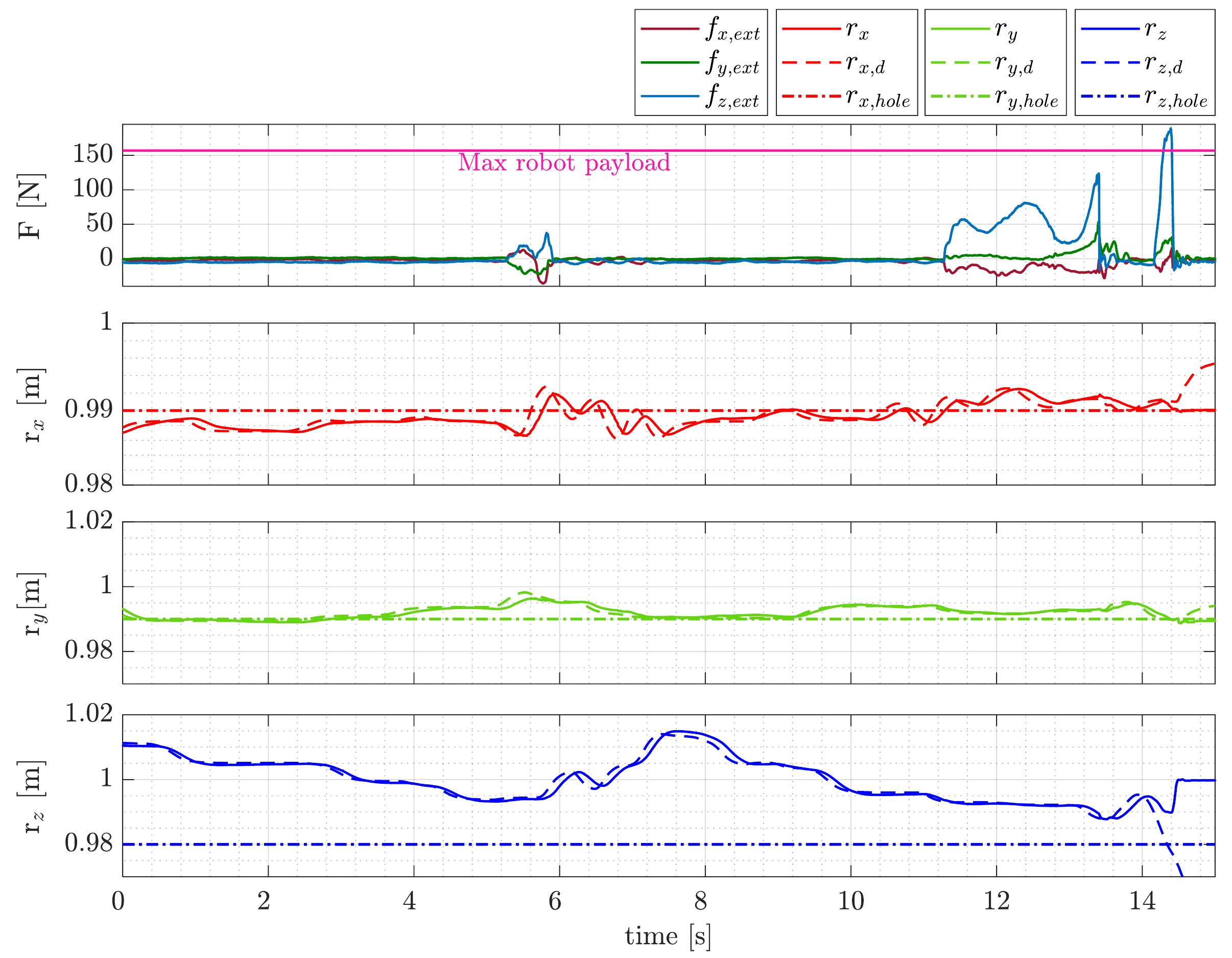}\\
    \hspace{0.4cm} (a) \hspace{8cm} (b)
    \caption{Results of the Constrained Interaction experiment. The outcomes for the peg-in-hole task are reported for MOCA-MAN and Kairos-MAN on the left side and right side, respectively. The plot on the top depicts the interaction forces of the end-effector with the environment and the maximum robot's payload. The other three plots report the x, y, and z coordinates of the desired (dashed line) and current (continuous line) end-effector position and the hole position (dotted line).}
    \label{fig:peginhole_results}
\end{figure*}


\subsection{Supplementary Experiments}
\label{subsec:Platform_Analysis}

\subsubsection{Constrained Interaction}
This experiment evaluates the interaction capabilities of MOCA-MAN and Kairos-MAN during co-assembly tasks. As a representative example, a collaborative peg-in-hole task was chosen, where a person guided the robot through the admittance interface to insert and remove a peg from a hole. In Fig.~\ref{fig:peginhole} the experimental setup is depicted for MOCA-MAN (left side) and Kairos-MAN (right side). In the experiment, the diameter of the peg is $33mm$ and the diameter of the hole is $34.5mm$.

In Fig.~\ref{fig:peginhole_results}a and~\ref{fig:peginhole_results}b, the results obtained for MOCA-MAN and Kairos-MAN are shown, respectively. From top to bottom, the two plots report the interaction forces of the end-effector with the environment ($\boldsymbol{f}_{ext}$), the end-effector current ($\boldsymbol{r}$) and desired  ($\boldsymbol{r}_d$, obtained from the admittance interface) positions and the position of the hole ($\boldsymbol{r}_{hole}$). In addition, the horizontal line in the force plot identifies the maximum payload of each robot arm, according to their data-sheet.  

The results for Kairos-MAN demonstrate that the system could not complete the task. A slight mismatch between reference and current position due to the interaction with the environment caused high forces, resulting in task failure. Consequently, the plot shows that the human could not insert the peg into the hole since the interaction force overcame the robot's maximum payload. In contrast, MOCA-MAN could complete the task smoothly. In this case, even though the interaction with the environment caused a mismatch between current and desired position, the forces that arose were small thanks to the whole-body Cartesian controller's impedance parameters, which allowed a compliant robot behavior.

\begin{figure*}
    \centering
    \includegraphics[width=1\textwidth]{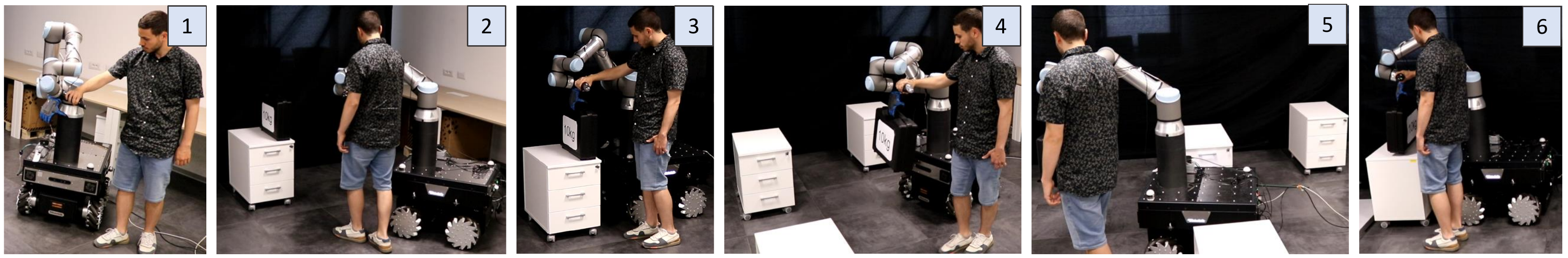}
    \caption{Excerpts of the main steps of the Long Distance Load Carrying task are depicted: the human-robot system 1, 2) approaches the box, 3) picks the box, 4, 5) carries the box to the drop location and 6) places the box.}
    \label{fig:loadandcommandsdemo}
\end{figure*}

\begin{figure*}
    \centering
    \includegraphics[width=0.9\columnwidth]{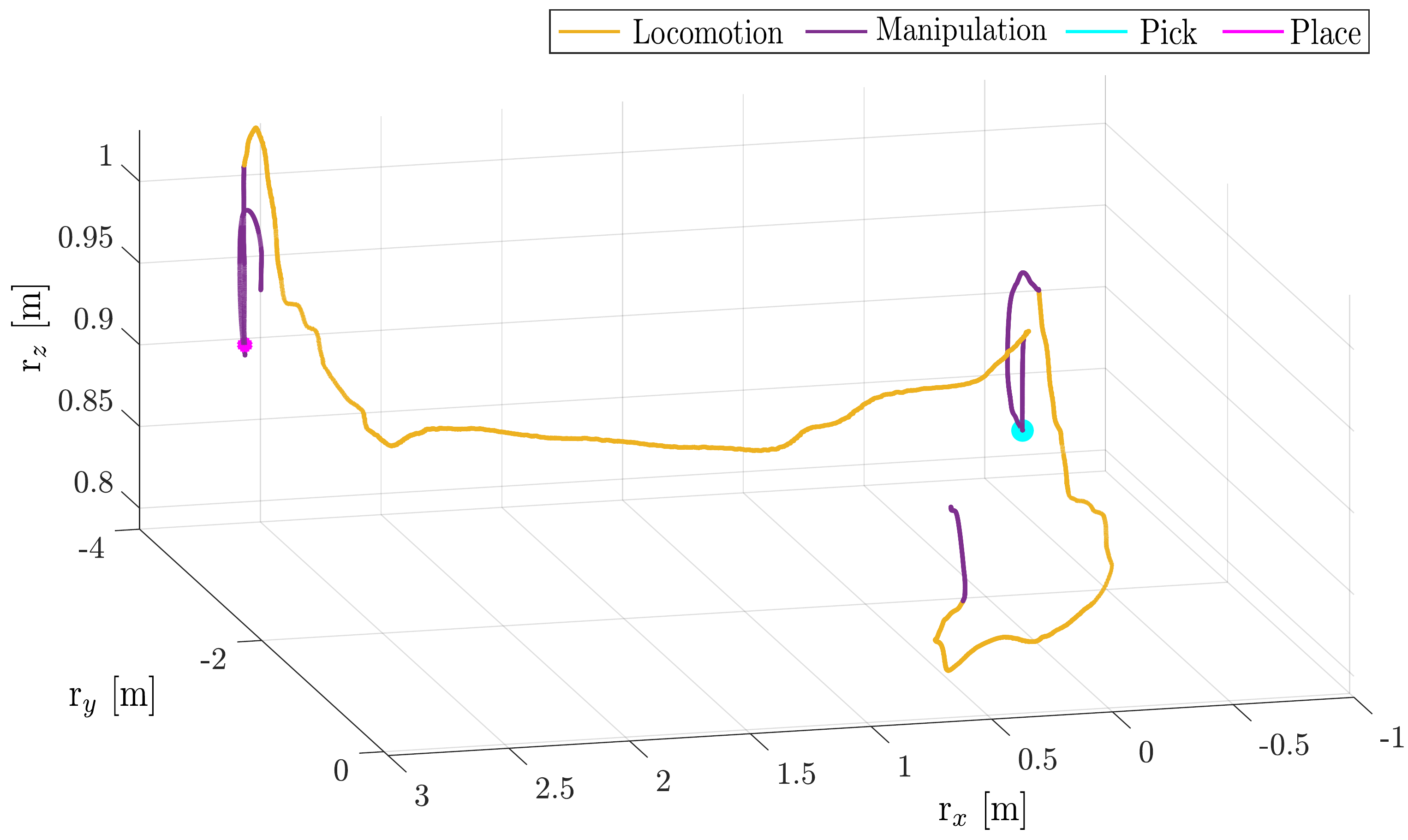}
    \hspace{0.2cm}
    \includegraphics[width=0.95\columnwidth]{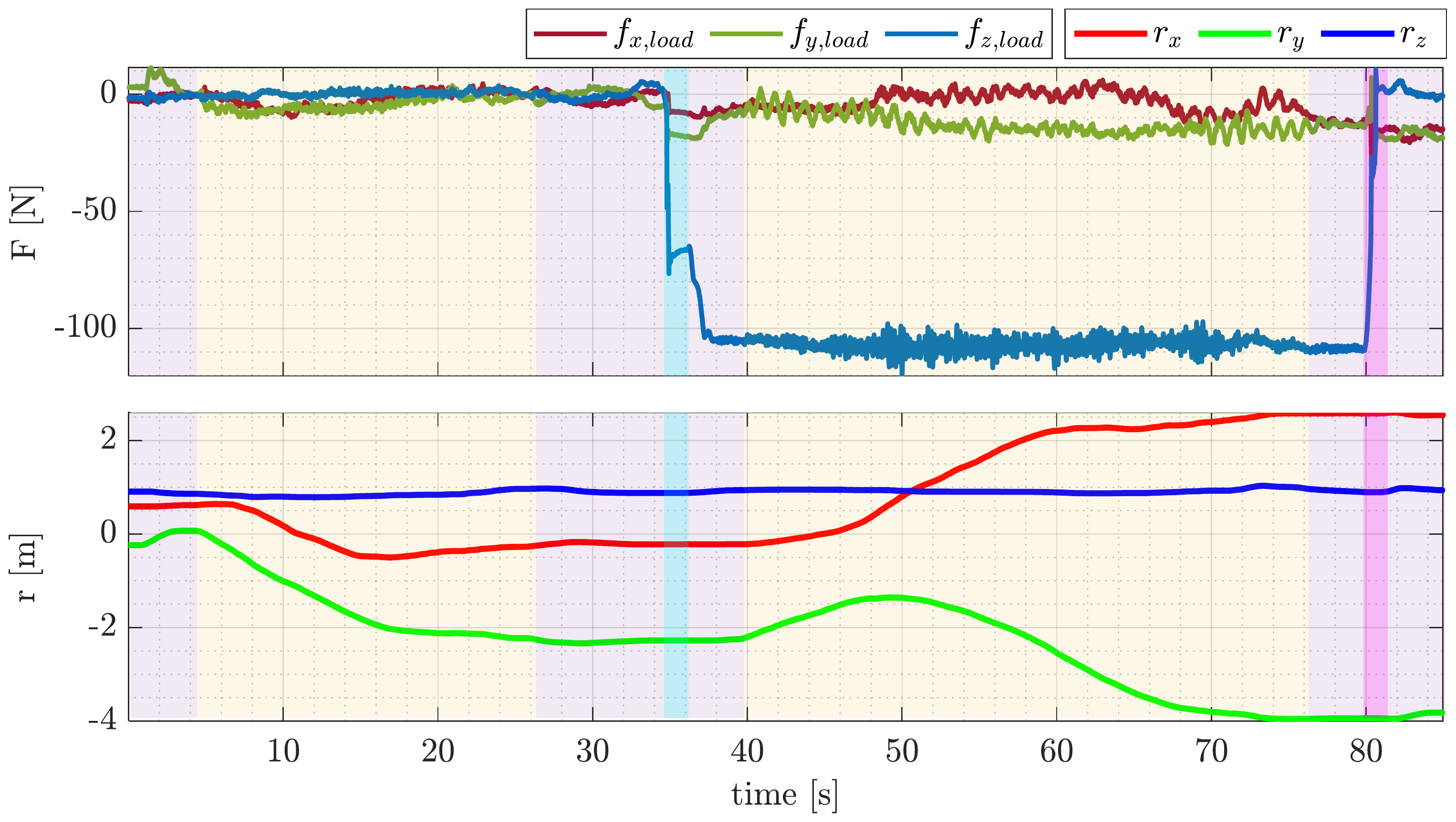}\\
    \hspace{0.4cm} (a) \hspace{8cm} (b)
    \caption{Results of the Long Distance Load Carrying experiment. On the left-side plot, a 3D representation of the complete end-effector path followed during the experiment. On the right-side plots, from the top to the bottom: The external forces applied at the end-effector of the robot, and the end-effector positions in the world reference frame.}
    \label{fig:loadandcommandsexperiment}
\end{figure*}

\subsubsection{Long Distance Load Carrying}

This experiment consisted of a collaborative pick-and-place of a heavy load (10kg) over a large workspace in the presence of static obstacles (see Fig.~\ref{fig:loadandcommandsdemo}). 

Fig.~\ref{fig:loadandcommandsexperiment} depicts the results of the experiment conducted using Kairos-MAN (since MOCA has a 3kg payload, it cannot perform this operation). In particular, on the left side, the 3D trajectory of the end-effector position during the whole task is shown, while on the right side, end-effector positions $\boldsymbol{r}$ and external forces $\boldsymbol{f}_{load}$ measured by the end-effector F/T sensor are reported as a function of time. The violet and yellow colors (line -- left-side plot, and background -- right-side plot) identify manipulation and locomotion phases, respectively. The light blue and purple (markers  -- left-side plot, and backgrounds -- right-side plot) determine the points where pick and place occurred, respectively.
The different steps of the task can be discerned in the right-side graph: at the beginning, no load was lifted, and the F/T sensor measured only the interaction forces between human and robot ($0s-35s$), then the load was grasped and lifted, resulting in $f_{z,load}$ reaching approximately $-100N$ ($35s-38s$). After that, the object was carried ($38s-80s$) and placed at the final location ($80s-82s$). Using the admittance-type interface, the operator successfully avoided the static obstacles and accomplished the task.


\section{Discussion}
\label{sec:discussion}

The experimental results highlighted some fundamental differences between the two implementations of the SUPER-MAN framework developed in this article. Thanks to the whole-body Cartesian impedance controller, MOCA-MAN features a compliant behavior that allows it to accomplish interactive tasks accommodating environment uncertainties and user inaccuracies. Conversely, the Kairos-MAN system does not succeed in these tasks since even minor errors can lead to considerable interaction forces, hence failing, causing a safety stop due to exceeded maximum payload.
On the other hand, the range of tasks that MOCA-MAN can cover is limited by its 3kg payload, while Kairos-MAN with a payload of 16kg can relieve workers from a wide range of physically demanding tasks in industrial environments.

Quantitative and qualitative results obtained from the user studies were presented in Section~\ref{subsec:User_Study}. The average muscles activity patterns shows that both platforms have a high potential in reducing the physical effort of the workers during industrial tasks. Indeed, in both user studies, the effort index improved. As expected, the improvement is more evident in the overhead drilling task due to its higher physical demanding nature.  

Quantitative results also supported the framework's potential for improving precision and endurance. Even if the precision improvement in the 3D path tracking experiment was not statistically significant, we believe that the effect of fatigue will be much less evident in the HR case in comparison to H for longer periods of time. Conversely, the overhead drilling study shows a significant enhancement both in endurance and precision due to an instantaneous loading. Only three subjects out of 12 could complete the task without the robot assistance. Moreover, all the participants made less or the same number of errors when cooperating with the robot.

Despite the improvements in precision and endurance, a significant deterioration of the completion time has been exhibited in both studies, which might be relevant in highly productive industrial environments. From the study conducted here, it is still unclear if the deterioration in completion time can be compensated by enhancing endurance, precision, and workers' condition. Further and long-term studies are needed to answer this question, even if it is well-known that, work-related musculoskeletal disorders that can be generated due to unhealthy working conditions (e.g., loading in our case), are among the main causes of economic loss in industry. 

From a qualitative point of view, the perceived workload of the participants was, in general, better with the robot than without, even though not all the workload subscales of the NASA-TLX questionnaire obtained statistically significant results. It is worth noticing that none of the subscales registered a statistically significant higher workload in HR than in H.

In summary, the general framework presented here has shown significant potential to improve workers' conditions in industrial environments. The two implementations considered have complementary features. Overall, both MOCA-MAN and Kairos-MAN demonstrated comparable task-related performance and reduced human perceived and actual workload w.r.t. the case where the robot does not assist the human. This achievement was accomplished in two demanding tasks: an over-the-shoulder task with high physical requirements and a 3D path tracking task requiring high dexterity and manipulation skills.

In contrast to exoskeletons and SRLs, the technology presented here is not wearable, therefore having the prospective to overcome the drawbacks and limitations of wearable devices. For instance, it does not need to be adapted precisely to each worker who wishes to use it, and it does not need to apply forces on the worker to function. In addition, with a human-centric design of the framework presented, it can collaborate with humans without hindering their natural movement, reducing their discomfort, which is one of the most critical drivers of workers' acceptance. Another important advantage of the framework presented over wearable devices is that, the former can be easily reconfigured to perform industrial tasks autonomously, while the latter can be used only in combination with humans, which limits their range of applicability.

The mobile base used in this work features omni-directional wheels, which would not allow the cooperating dyad to move in rough terrains or to climb stairs. This issue can be addressed by the extending the SUPER-MAN concept to legged or flying robots, that would be more suitable for traversing complex terrains. In addition, hybrid robots, e.g. having wheels and legs, can exploit the advantages of multiple platform types. 

Last but not least, another big challenge is related to safety. Since these platforms come in contact with humans, safety issues can arise and potential dangerous situations should be investigated and handled.

Overall, we can affirm that the outcomes obtained from this study show the high potential of the system in reducing workers' effort, health risks and in enhancing performance in industrial tasks like overhead drilling and prolonged manipulations. Although the experiments were conducted in a laboratory setting, the system is ready to be tested in real industrial environments. In future work, we will test the applicability of the framework in this kind of settings in order to get insights on the challenges and related solutions of real scenarios.

\section{Conclusions}
\label{sec:conclusions}
This work presented a general framework, namely SUPER-MAN, for using a floating base robot to assist and augment humans while performing loco-manipulation industrial-like tasks. The relevant system, e.g., Exoskeletons, SRLs and Cobots, were analyzed and their challenges and main drawbacks were discussed. Two possible implementations of the SUPER-MAN framework, i.e. MOCA-MAN and Kairos-MAN, were proposed and tested. 
The potential of the SUPER-MAN approach was revealed during the execution of constrained interactive and long distance load-carrying tasks. The user acceptance of the two systems and their potential at relieving humans workload and at augmenting their capabilities were experimentally evaluated through quantitative and qualitative analysis of user studies involving 12 subjects. The user studies targeted the execution of a prolonged and precise task with a low/medium effort level, and an awkward posture and effort-demanding manipulation task, as common examples in industrial scenarios (e.g., automotive and warehouse shop-floors).   
Overall, the results showed that the developed framework can potentially improve workers' conditions in tasks having different requirements while maintaining a good level of task-related performance and obtaining positive outcomes for user acceptance. Some advantages and drawbacks w.r.t. the state-of-art technologies were also discussed.

Future work will focus on further exploring the potential of the SUPER-MAN framework by designing additional user studies. Indeed, evaluating the acceptance and the effects  of these technologies on real workers is of fundamental importance for determining their actual advantage. Besides, these studies will allow to better understand the challenges and limitations of the framework, and will ease the development of human-centric design of its three main blocks, namely the admittance-type interface, the HR interaction controller and the whole-body controller of the supernumerary body.



\appendix

\section{Mathematical Notation and Symbols}
\label{appendix:math_not_symb}
Main mathematical notation and symbols used throughout the text are defined in Table~\ref{tab:notation}.
\begin{table*}
    \caption{Mathematical Notation and Symbols}
    \label{tab:notation}
    \centering
    \begin{tabular}{m{2.9cm}|p{12cm}}
    \textbf{Symbol} & \textbf{Description} \\
    \hline
    $n_a$ & Arm DoFs \\ 
    $n_b$ & Mobile base DoFs \\ 
    $n$ & Whole body DoFs \\
    $\boldsymbol{q}, {\boldsymbol{\dot{q}}},{\boldsymbol{\ddot{q}}} \in \mathbb{R}^{n}$ & Current whole-body joint positions, velocities and accelerations \\
    $\boldsymbol{q}_a, {\boldsymbol{\dot{q}}}_a,{\boldsymbol{\ddot{q}}}_a \in \mathbb{R}^{n_a}$ & Current arm joint positions, velocities and accelerations \\
    $\boldsymbol{q}_b, {\boldsymbol{\dot{q}}}_b,{\boldsymbol{\ddot{q}}}_b \in \mathbb{R}^{n_b}$ & Current mobile base joint positions, velocities and accelerations \\
    $\boldsymbol{q}_d, {\boldsymbol{\dot{q}}}_d,{\boldsymbol{\ddot{q}}}_d \in \mathbb{R}^{n}$ & Desired whole-body joint positions, velocities and accelerations \\
    $\boldsymbol{M}_a(\boldsymbol{q}_a) \in \mathbb{R}^{n_a \times n_a}$ & Arm mass matrix \\
    $\boldsymbol{C}_a(\boldsymbol{q}_a,\boldsymbol{\dot{q}}_a) \in \mathbb{R}^{n_a \times n_a}$ & Arm Coriolis and centrifugal terms matrix \\
    $\boldsymbol{g}_a(\boldsymbol{q}_a) \in \mathbb{R}^{n_a}$ & Arm gravity vector \\
    $\boldsymbol{\tau}_{a,ext}, \boldsymbol{\tau}_{a} \in \mathbb{R}^{n_a}$ & Arm external and control torque vectors \\
    $\boldsymbol{M}_{v}, \boldsymbol{D}_{v} \in \mathbb{R}^{n_b \times n_b}$ & Diagonal positive definite virtual mass and damping matrix of the mobile base \\
    $\boldsymbol{\tau}_{v} \in \mathbb{R}^{3}$ & Virtual mobile base torques vector \\
    $\boldsymbol{M}_{adm}, \boldsymbol{D}_{adm} \in \mathbb{R}^{6 \times 6}$ & Diagonal positive definite admittance mass and damping matrix \\
    $\boldsymbol{x}_{d}, \boldsymbol{\dot{x}}_{d}, \boldsymbol{\ddot{x}}_{d} \in \mathbb{R}^{6}$ & Desired end-effector pose, twist and acceleration \\
    $\boldsymbol{r}_d, \boldsymbol{\dot{r}}_d, \boldsymbol{\ddot{r}}_d \in \mathbb{R}^{3}$ & Desired end-effector position, linear velocity and linear acceleration \\
    $\boldsymbol{\theta}_d, \boldsymbol{\omega}_d, \boldsymbol{\dot{\omega}}_d \in \mathbb{R}^{3}$ & Desired end-effector orientation, angular velocity and angular acceleration \\
    $\boldsymbol{x}, \boldsymbol{\dot{x}}, \boldsymbol{\ddot{x}} \in \mathbb{R}^{6}$ & Current end-effector pose, twist and acceleration \\
    $\boldsymbol{r}, \boldsymbol{\dot{r}}, \boldsymbol{\ddot{r}} \in \mathbb{R}^{3}$ & Current end-effector position, linear velocity and linear acceleration \\
    $\boldsymbol{\theta}, \boldsymbol{\omega}, \boldsymbol{\dot{\omega}} \in \mathbb{R}^{3}$ & Current end-effector orientation, angular velocity and angular acceleration \\
    $\hat{\boldsymbol{\lambda}}_h \in \mathbb{R}^{6}$ & Measured human wrench \\
    $\hat{\boldsymbol{f}}_h, \hat{\boldsymbol{\tau}}_h \in \mathbb{R}^{3}$ & Measured human force and torque \\
    $\mathcal{N}(\cdot)$ & Null-space operator \\
    $det(\cdot)$ & Determinant operator \\
    $\boldsymbol{J}(\boldsymbol{q})\in \mathbb{R}^{6 \times n}$ & Whole body geometric Jacobian \\
    $\boldsymbol{F} \in \mathbb{R}^6$ & Cartesian generalized forces \\
    $\bar{\boldsymbol{J}} \in \mathbb{R}^{n \times 6}$ & Dynamically consistent Jacobian \\
    $\boldsymbol{\Lambda}, \boldsymbol{\Lambda_{W}} \in \mathbb{R}^{6 \times 6}$ & Unweighted and weighted Cartesian Inertia \\
    $\boldsymbol{W} \in \mathbb{R}^{n \times n}$ & Positive definite weighting matrix \\
    $\eta_A, \eta_B \in \mathbb{R}_{>0}$ & Loco-manipulation gains \\
    $\boldsymbol{D}_d, \boldsymbol{K}_d \in \mathbb{R}^{6 \times 6}$ & Desired damping and stiffness of Cartesian impedance controller \\
    $\boldsymbol{D}_0, \boldsymbol{K}_0 \in \mathbb{R}^{n \times n}$ & Null-space damping and stiffness of Cartesian impedance controller \\
    $\boldsymbol{q}_{pref} \in \mathbb{R}^6$ & Preferred configuration vector \\
    $\boldsymbol{\tau}_0 \in \mathbb{R}^n$ & Null-space Torque \\
    $\boldsymbol{H} \in \mathbb{R}^{n \times n}$ & Diagonal positive definite controller weighting matrix \\
    ${\boldsymbol{\dot{q}}}_{d,1},{\boldsymbol{\dot{q}}}_{d,2} \in \mathbb{R}^{n}$ & Desired first and second priority joint velocities \\
    $\boldsymbol{K} \in \mathbb{R}^{6 \times 6}$ & Pose feedback gain matrix \\
    $\boldsymbol{W}_1 \in \mathbb{R}^{6 \times 6}$ & Tracking error diagonal weight matrix \\
    $\boldsymbol{W}_2 \in \mathbb{R}^{n \times n}$ & Regularization diagonal weight matrix \\
    $w_a, w_b \in \mathbb{R}$ & Arm and mobile base weights \\
    $\mathcal{L}_1(\boldsymbol{\dot{q}}), \mathcal{L}_2(\boldsymbol{q})$ & First and Second priority cost functions \\
    $w(\boldsymbol{q}_a) \in \mathbb{R}$ & Arm manipulability index \\
    $w_t \in \mathbb{R}$ & Arm manipulability minimum threshold \\
    $\boldsymbol{J}_a(\boldsymbol{q}_a) \in \mathbb{R}^{6 \times n_a}$ & Arm geometric Jacobian \\
    $k \in \mathbb{R}$ & Damping factor \\
    $k_0 \in \mathbb{R}$ & Damping factor parameter \\
    $k_i \in \mathbb{R}$ & Priority mode gain \\
    $\boldsymbol{G}_i \in \mathbb{R}^{n \times n}$ & Secondary task diagonal weight matrix \\
    $\xi \in \mathbb{R}$ & Damping coefficient \\
    \hline
    \end{tabular}
\end{table*}

\section{Choice of Controller Parameters}
\label{appendix:controller_params}
The following appendix lists the specific values of the parameters utilized in the robot controllers. Please note that any blank entries in the matrices should be interpreted as being equal to zero.

\subsection{MOCA Whole-Body Controller}
Cartesian stiffness and damping in (\ref{eq:cartesian_impedance}) are experimentally chosen in order to find a trade-off between compliance and tracking accuracy while guaranteeing a stable behavior. Their values are set to 
\begin{equation*}
\begin{aligned}
\boldsymbol{K}_d &= \begin{bmatrix}
 500 \boldsymbol{I}_3 & \\
  & 30 \boldsymbol{I}_3
\end{bmatrix}, \\
\boldsymbol{D}_d &= 2\xi \boldsymbol{K}_d^{\frac{1}{2}},
\end{aligned}
\end{equation*}
with $\xi=0.7$. 

Nullspace stiffness and damping in (\ref{eq:nullspace_impedance}) have different values in the two priority modes, manipulation and locomotion. In manipulation, $\boldsymbol{K}_0$ was chosen as
\begin{equation*}
\boldsymbol{K}_0 = \begin{bmatrix}
 \boldsymbol{0}_3 & \\
  & 5 \boldsymbol{I}_7
\end{bmatrix}.
\end{equation*}
The values are low in order to avoid undesired movements of the mobile base while allowing high joints compliance. In locomotion, higher values are chosen 
\begin{equation*}
\boldsymbol{K}_0 = \begin{bmatrix}
 \boldsymbol{0}_3 & \\
  & 50 \boldsymbol{I}_7
\end{bmatrix},
\end{equation*}
so that the base follows arm's movements. In both cases $\boldsymbol{D}_0=2\xi\boldsymbol{K}_0^{\frac{1}{2}}$. 

The virtual mass and damping in (\ref{eq:admittance_eq_base}) are selected in order to find an acceptable behavior for the mobile base: the movements should be neither too jerky nor too slow. Hence, these were set to %
\begin{equation*}
\boldsymbol{M}_v = \begin{bmatrix}
 105 & & \\
 & 105 & \\
 & & 21
\end{bmatrix},
\end{equation*}
where $105kg$ is the actual mass of the mobile base, and $\boldsymbol{D}_v=10\boldsymbol{M}_v$. Finally, in order to assign most of the movement to the arm during manipulation, $\eta_B=5$ and $\eta_A=1$, vice-versa in locomotion $\eta_B=1$ and $\eta_A=6$. Note that, in locomotion most of the movement is assigned to the mobile base also thanks to the choice of $\boldsymbol{K}_0$.

\subsection{Kairos Whole-Body Controller}
In order to allow a good tracking performance of the desired end-effector movement ($\boldsymbol{x}_d$ and $\boldsymbol{\dot{x}}_d$) while avoiding to generate too high joint velocities (especially when close to singularity), $\boldsymbol{K}$, $\boldsymbol{W}_1$ are experimentally tuned.

The final values obtained are 
\begin{equation*}
\begin{aligned}
\boldsymbol{K} &= \begin{bmatrix}
 0.5 \boldsymbol{I}_3 & &\\
 &  0.05 & \\
 & & 0.01 \boldsymbol{I}_2
\end{bmatrix}, \\
\boldsymbol{W}_1 &= 100\cdot \begin{bmatrix}
10 \boldsymbol{I}_3 &\\
 &  5 \boldsymbol{I}_3\\
\end{bmatrix}.
\end{aligned}
\end{equation*}
In the same way, $k_0=2$ and $w_t=0.001$ have been experimentally determined.

The desired motion at the end-effector is distributed between the mobile base and the arm depending on the priority mode (manipulation or locomotion) through heuristic tuning of $w_b$, $w_a$ and $k_i$. These are set respectively to $10$, $0.5$ and $1$ in locomotion and $100$, $1$ and $0$ in manipulation.

\subsection{Admittance Controller}
For the experiments conducted in this work, different admittance parameters in (\ref{eq:admitcontrollaw}) have been selected according to the type of task to be accomplished.

\subsubsection{User Studies}
A relatively low level of admittance is used to set a trade-off between human effort and movement accuracy (i.e., controllability of the robot movements through the admittance interface). Hence, the parameters selected are 
\begin{equation*}
\begin{aligned}
\boldsymbol{M}_{adm} &= \begin{bmatrix}
 6 \boldsymbol{I}_3 &\\
 &  \boldsymbol{I}_3 \\
\end{bmatrix}, \\
\boldsymbol{D}_{adm} &= \begin{bmatrix}
20 \boldsymbol{I}_3 &\\
 &  1.5 \boldsymbol{I}_3\\
\end{bmatrix}.
\end{aligned}
\end{equation*}
\subsubsection{Supplementary Experiments}
In order to keep a stable behavior while compromising accuracy and transparency, the values obtained experimentally are 
\begin{equation*}
\begin{aligned}
\boldsymbol{M}_{adm} &= \begin{bmatrix}
 3 \boldsymbol{I}_3 &\\
 &  \boldsymbol{I}_3 \\
\end{bmatrix}, \\
\boldsymbol{D}_{adm} &= \begin{bmatrix}
20 \boldsymbol{I}_3 &\\
 &  1.5 \boldsymbol{I}_3\\
\end{bmatrix}.
\end{aligned}
\end{equation*}

\section*{Acknowledgments}
This work was supported in part by the European Research Council's (ERC) starting grant Ergo-Lean (GA 850932) and in part by the European Union’s Horizon 2020 research and innovation program CONCERT (GA 101016007).

\bibliographystyle{elsarticle-num} 
\bibliography{biblio.bib}





\end{sloppypar}

\end{document}